\documentclass{article}

\PassOptionsToPackage{numbers, compress}{natbib}

\usepackage[preprint]{neurips}

\usepackage[utf8]{inputenc} %
\usepackage[T1]{fontenc}    %
\usepackage{hyperref}       %
\usepackage{url}            %
\usepackage{booktabs}       %
\usepackage{amsfonts,amsmath,amssymb,amsthm}       %
\usepackage{nicefrac}       %
\usepackage{microtype}      %
\usepackage{xcolor}         %
\usepackage{graphicx}

\title{Vibe Patenting: Evaluating LLM Judges for Professional Patent-Drafting Agents}

\author{%
  Toshiaki Koike-Akino, Vladislav Blaykhman, Ye Wang, Jing Liu, Gene V. Vinokur\\
  Mitsubishi Electric Research Laboratories (MERL)\\
  201 Broadway, Cambrdige, MA 02139 \\
  \texttt{\{koike, blyakhman, yewang, jiliu, vinokur\}@merl.com} \\
}

\begin{document}

\maketitle

\begin{abstract}
LLM judges are increasingly used to evaluate and improve AI-generated outputs, yet their reliability for complex professional work remains unclear. 
We study this problem through \textbf{Vibe Patenting}, an end-to-end patent-drafting testbed for AI-agent evaluation. 
A separately-invoked LLM judge evaluates generated patent drafts and provides structured feedback for iterative revision. 
Across multiple inventions and drafting-agent configurations, judge-guided revision consistently improves judge-assessed quality, while unguided revision tends to saturate. 
Notably, iterative judge feedback enables a low-reasoning agent to approach the performance of a substantially more expensive high-reasoning agent. 
Stronger models and increased reasoning generally improve judge-assessed drafting quality, while domain-specific agentic workflows provide further gains. 
We validate the judge against independent evaluation by a professional patent attorney and find meaningful but strongly metric-dependent agreement and systematic calibration differences. 
These results highlight both the utility and limitations of LLM judges as evaluators and optimization signals for complex professional workflows.
\end{abstract}

\section{Introduction}
\label{sec:introduction}

Large language models (LLMs) are increasingly used not only to generate content, but also to \emph{evaluate} it. 
\textbf{LLM-as-a-judge} methods provide scalable evaluation of open-ended outputs for which conventional automatic metrics are inadequate~\cite{zheng2023judging,liu2023geval}. 
As AI systems become more agentic, however, the role of the judge is expanding from an offline evaluation tool to an active component of the agentic pipeline: a judge can critique an output, provide structured feedback, and guide subsequent revision. 
\emph{Can LLM judges provide reliable evaluation and useful guidance for complex professional work?}

We investigate this question through \textbf{Vibe Patenting}, the automated transformation of technical materials into professional patent drafts with minimal human drafting effort. 
Patent drafting provides a challenging
evaluation setting because quality is multidimensional: a complete draft must exhibit strong claims, sufficient technical disclosure,
coherent figures, prosecution resilience, and overall filing readiness.
Meaningful assessment ordinarily requires specialized professional expertise.

We compare patent drafts produced by a skilled human drafter and LLM systems
spanning different model generations, reasoning budgets, and agent
architectures, using an independent LLM judge for patent quality assurance (QA). 
We first
examine how judge-assessed professional quality varies across these drafting
approaches. We then place the judge inside an iterative revision loop and
compare judge-guided refinement with generic revision. Guided revision
continues to improve judge-assessed quality as unguided revision tends to
saturate; notably, iterative feedback enables a low-reasoning agent to
approach the performance of a substantially more expensive high-reasoning agent.

Finally, we separately validate the LLM judge against evaluations from a
professional patent attorney. Agreement is strongly metric dependent: some
quality dimensions exhibit meaningful ranking agreement despite systematic
score bias, whereas others show substantially weaker correlation. Thus,
improvement under an automated judge does not by itself establish improvement
under professional expert judgment. Our results position patent drafting as a
practical testbed for studying the \emph{validity, calibration, and
feedback-loop behavior} of LLM judges in professional agentic workflows.

\begin{figure*}[t]
    \centering
    \includegraphics[width=\textwidth,trim=0 100 0 60,clip]{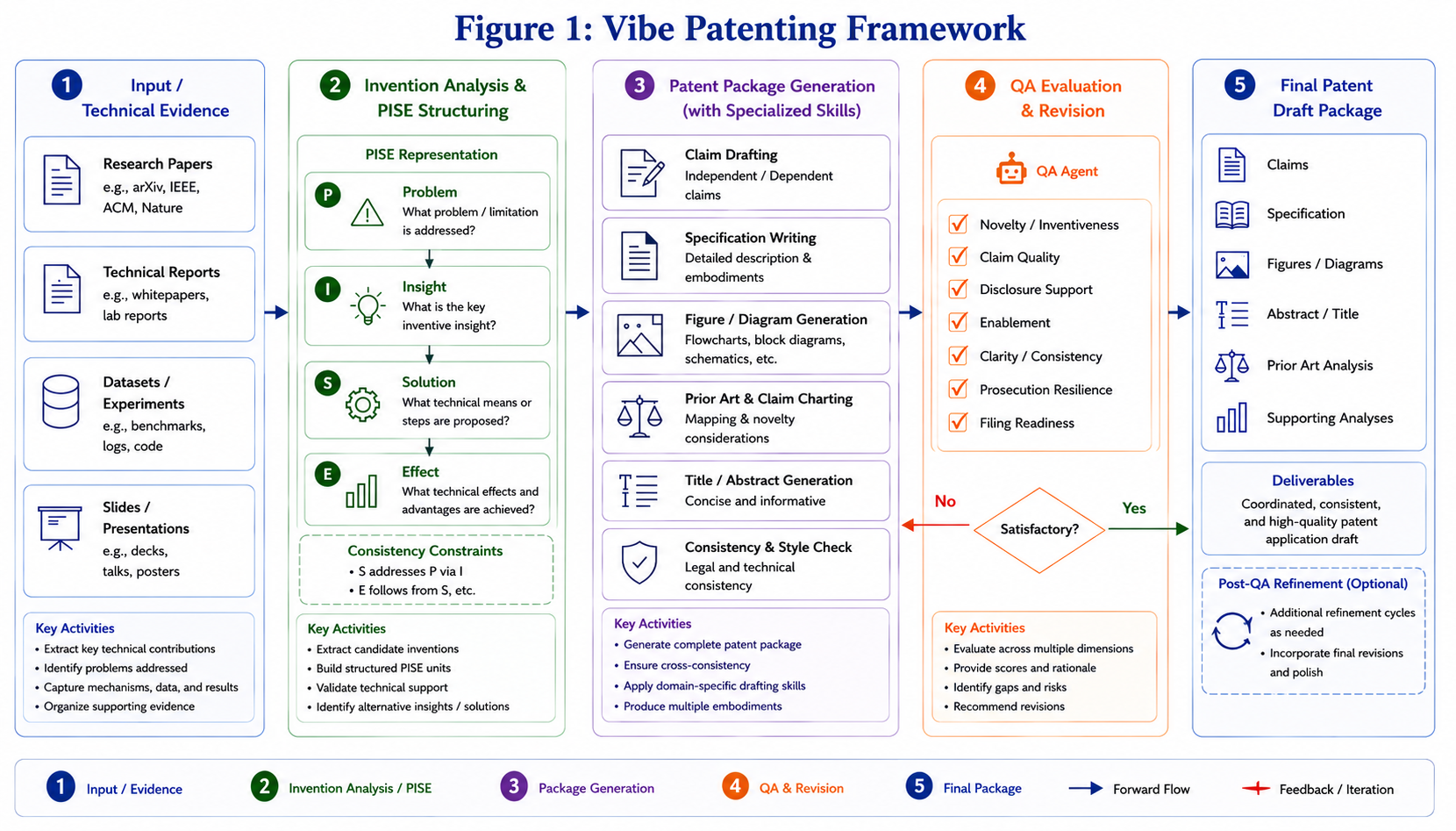}
    \caption{
    Vibe Patenting framework. AI agent first analyzes technical evidence into a shared representation. 
    Patent-specific skills then generate coordinated claims, specification, figures, and supporting analyses. 
    QA judge agent evaluates the resulting package and can initiate revision cycle.
    }
    \label{fig:vibe-framework}
\end{figure*}

\section{Vibe Patenting: Professional Evaluation Testbed}

We use \emph{Vibe Patenting} as a testbed for studying LLM judges in a complex professional workflow. Given technical source material $X$, the AI drafting system produces a coordinated patent package
$D=\{C,S,F,A\}$,
where $C$ denotes claims, $S$ the written specification, $F$ patent figures, and $A$ supporting analyses. Unlike ordinary text generation, these components are mutually constrained: claim scope should be supported by the specification, figures should reflect the disclosed embodiments, and terminology and technical content should remain consistent across the package. Figure~\ref{fig:vibe-framework} summarizes the overall workflow (See Appendix \ref{app:framework}).

Our AI agent can incorporate professional skills and domain expertise analogous to those used by a patent attorney in drafting.
Specifically, the drafting pipeline first analyzes the source material, and structures the candidate invention using a Problem--Insight--Solution--Effect (PISE) representation (see Appendix~\ref{app:pise}). 
PISE captures the technical problem, the inventive insight, the proposed solution, and its resulting technical effects, and provides a shared representation for downstream drafting. 
Patent-oriented skills then use it to generate coordinated claims, specification, embodiments, figures, and supporting analysis. 
We evaluate several drafting configurations: general-purpose chat AI; meta agent; and domain-specialized custom agent systems (see Appendix \ref{app:framework}).

\paragraph{LLM judge.}
A separate patent-QA LLM judge evaluates each completed draft along five professional quality dimensions: \emph{Prosecution Resilience}, \emph{Claim Strength}, \emph{Disclosure Strength}, \emph{Figure Quality}, and \emph{Filing Readiness}. 
The QA judge uses rubric-coordinated scoring from 1 to 10 and provides structured critique and revision recommendations (see Appendix~\ref{app:judge}).
Each draft is evaluated independently five times, and we average the repeated scores for each dimension; the overall score is the mean across the five dimensions.

\paragraph{Judge-guided revision.}
The QA judge can initiate an iterative optimization loop. 
At revision round $t$, the judge evaluates draft $D_t$ to produce a structured QA report
$
Q_t=f_{\mathrm{judge}}(D_t).
$
The drafting AI agent then receives the current draft and the QA report to produce a revision
$
D_{t+1}=f_{\mathrm{rev}}(X,D_t,Q_t)$.
We compare this judge-guided procedure with a generic revision control in which the same drafting AI agent is asked to improve $D_t$ without access to $Q_t$ as $
D_{t+1}'=f_{\mathrm{rev}}(X,D_t)$. 
See Appendix~\ref{app:qa-loop}.

Finally, we distinguish two human roles in our experiments. 
A \emph{skilled human drafter} (having hundreds of filed coinventions) provides a human-generation baseline, whereas a separate \emph{professional patent attorney evaluator} independently scores available drafts for validating the LLM judge. The attorney evaluation is therefore used as expert reference judgment, not as the human drafting baseline.

\section{Experiments and Results}
\label{sec:experiments}

\paragraph{Experimental setup.}
We evaluate more than one hundred patent drafts generated from ten scientific
reports spanning multiple technical domains. 
Drafting configurations include
GPT-5.6 Sol chat mode with different reasoning levels (Instant, Extra-High, Pro, etc.), GPT-5.4--5.6 agent modes,
a domain-specialized custom patent agent, and a skilled human drafter baseline.
Each completed draft is independently evaluated five
times by the LLM-based QA judge. The judge assigns integer scores from 1--10
for five dimensions; we report
their mean across dimensions as the overall score.

\subsection{Scaling Professional Patent-Drafting Agents}
\label{sec:scaling-results}

\begin{figure}[t]
    \centering
    \includegraphics[width=0.55\linewidth]
    {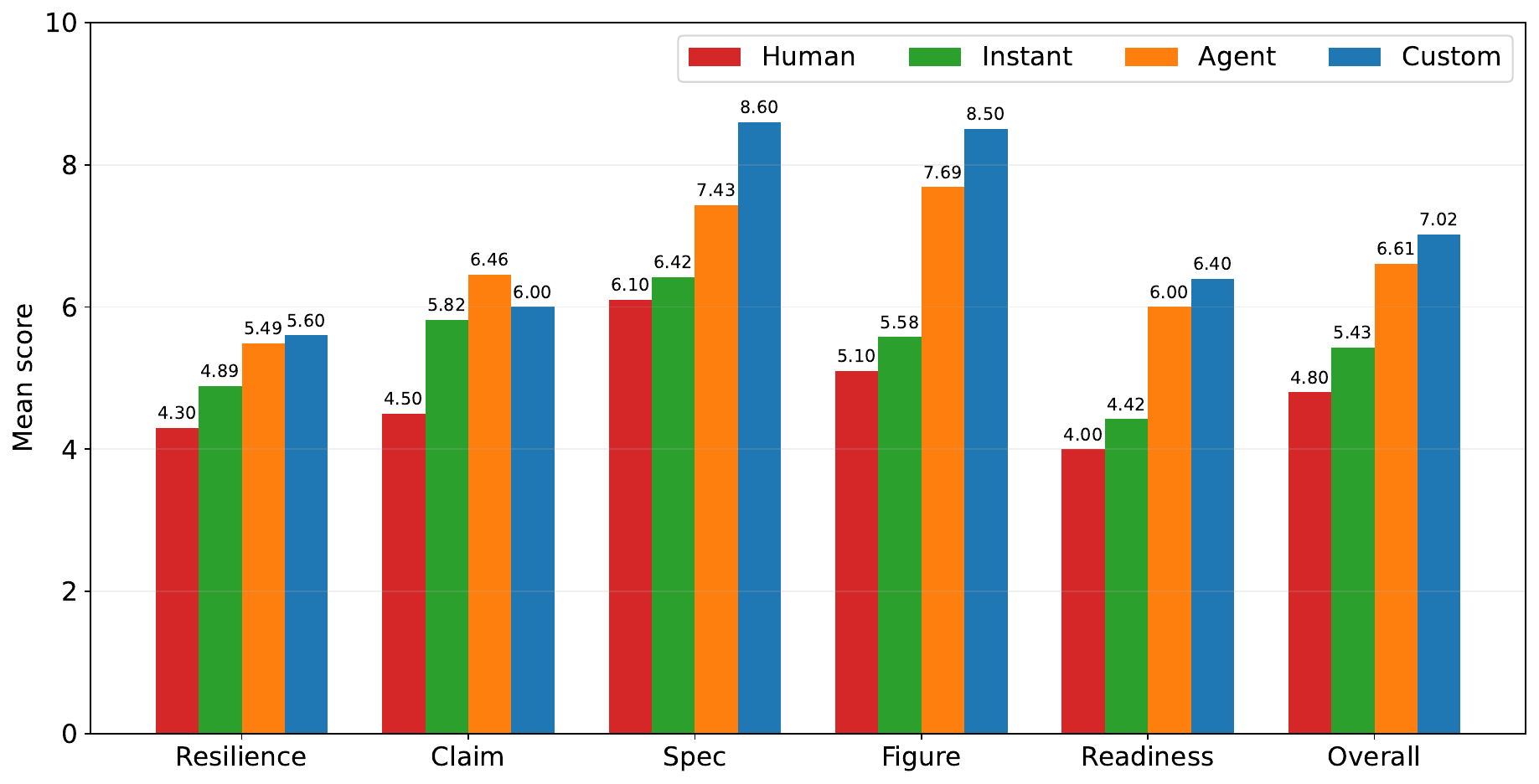}
    \hfill
    \includegraphics[width=0.44\linewidth]
    {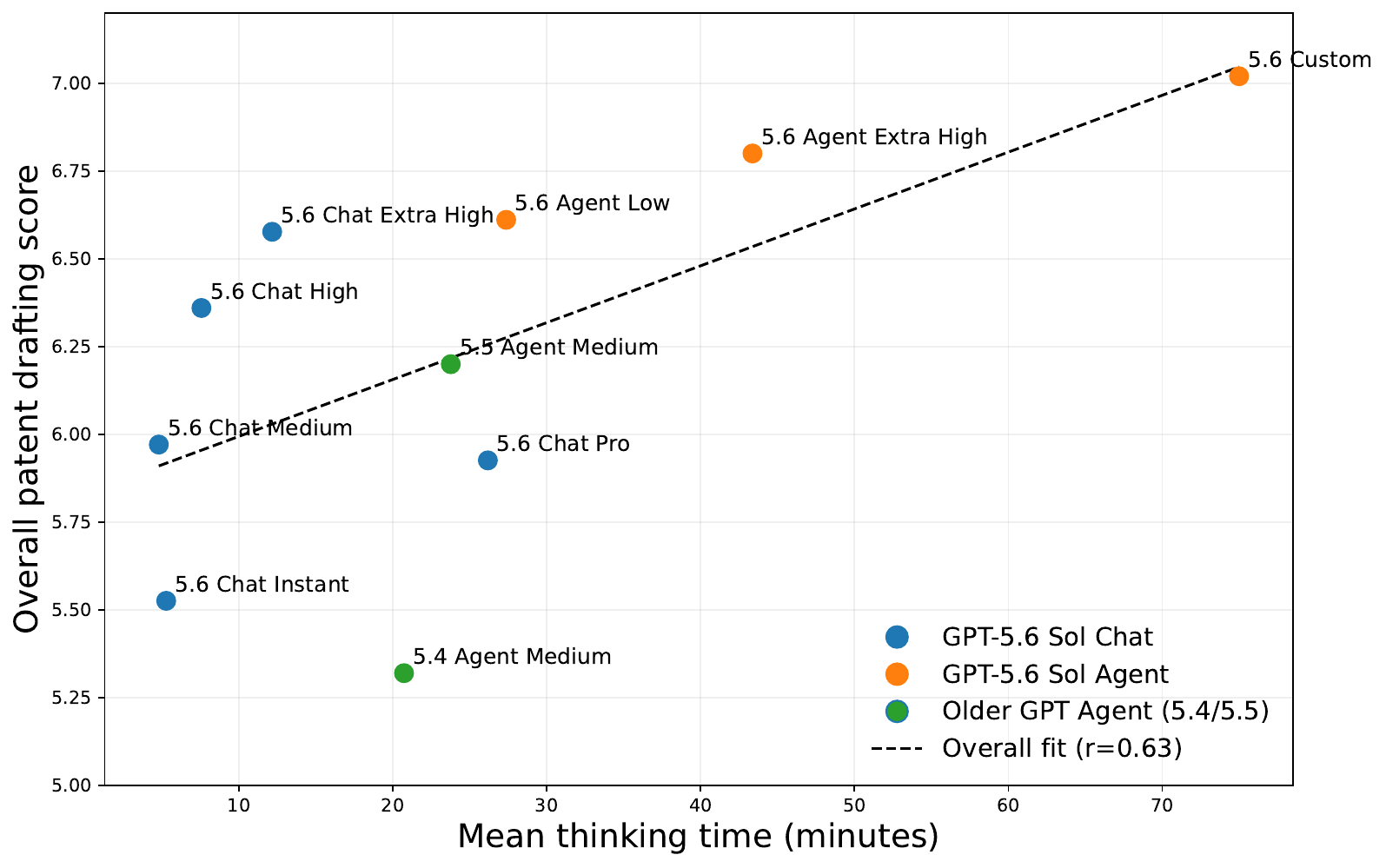}
    \caption{
    \textbf{Left:} Judge-assessed patent quality across drafting
    configurations. 
    \textbf{Right:} Overall judge score versus mean thinking time across
    model, reasoning, and agent configurations.
    }
    \label{fig:scaling}
\end{figure}

Figure~\ref{fig:scaling} (Left) compares drafting approaches under GPT-5.6 Sol.
Agentic scaffolding improves upon zero-shot generation, while the
domain-specialized custom agent achieves the highest overall judge score.
Under the LLM judge, most AI configurations receive scores higher than the skilled-drafter baseline, particularly for Disclosure and Figure Quality.
Importantly, the human
drafter is a \emph{generation baseline} and is distinct from the professional
patent attorney used later for independent expert evaluation.

Figure~\ref{fig:scaling} (Right) further compares overall judge score with mean
thinking time across LLM model, reasoning, and agent configurations. Quality
generally increases with computation (Pearson $r=0.63$), but execution time alone does not
determine performance: configurations with similar thinking times can achieve
substantially different scores. This suggests that how computation is
structured---through model capability, reasoning, and agentic
scaffolding---matters in addition to its amount.

\subsection{Judge-Guided Iterative Revision}
\label{sec:revision-results}

We next test whether the LLM judge can serve as an optimization signal rather than only an offline evaluator. 
We compare a $2\times2$ design consisting of Instant and Extra-High reasoning, each with
either structured QA feedback or a generic revision without
access to the QA report.

\begin{figure*}[t]
    \centering
    \begin{minipage}[t]{0.47\textwidth}
        \centering
        \includegraphics[width=\linewidth]
        {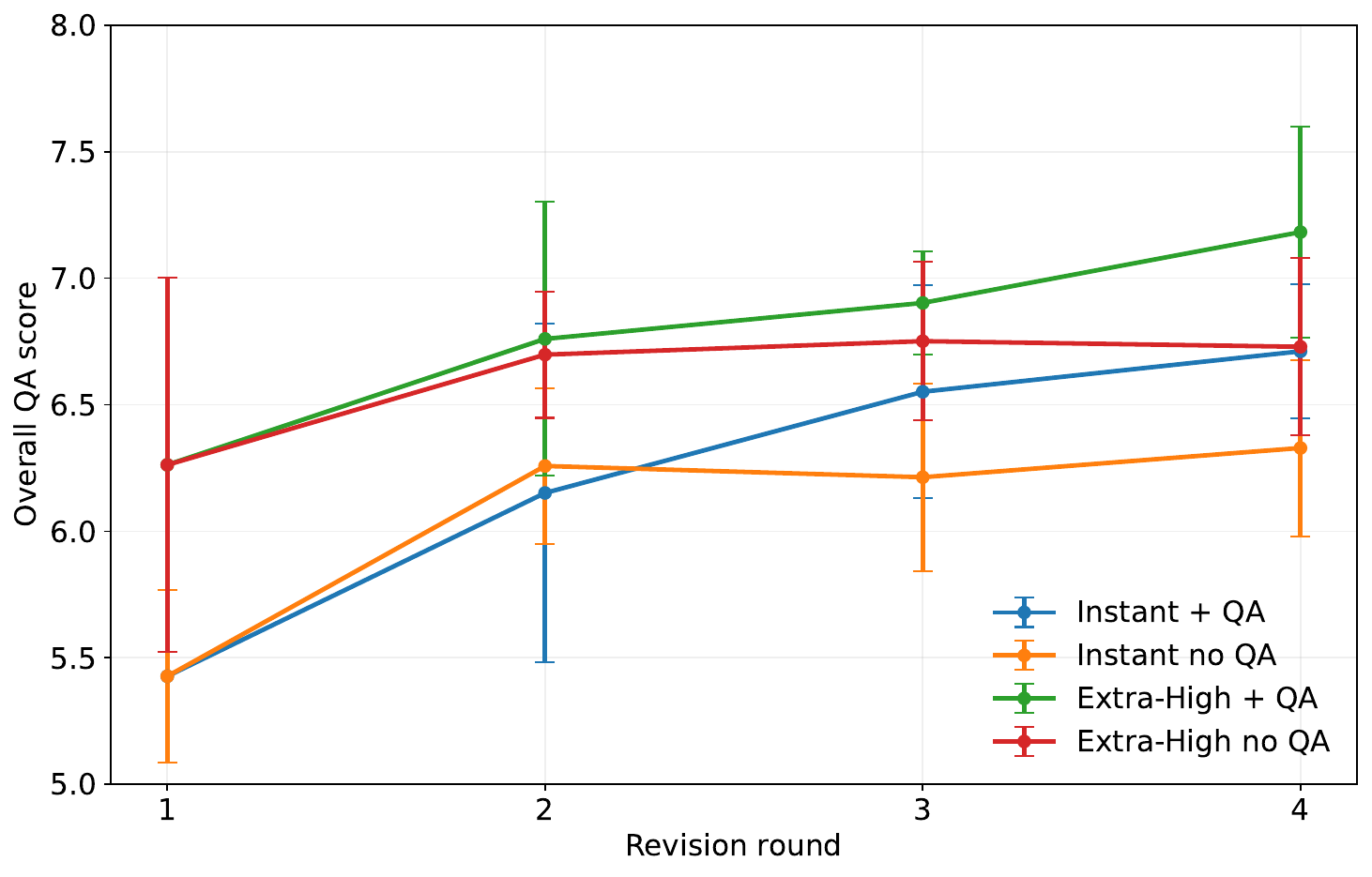}
        \vspace{-1mm}
        \textbf{(a)} Judge-guided iterative revision
    \end{minipage}
    \hfill
    \begin{minipage}[t]{0.52\textwidth}
        \centering
        \includegraphics[width=\linewidth]
        {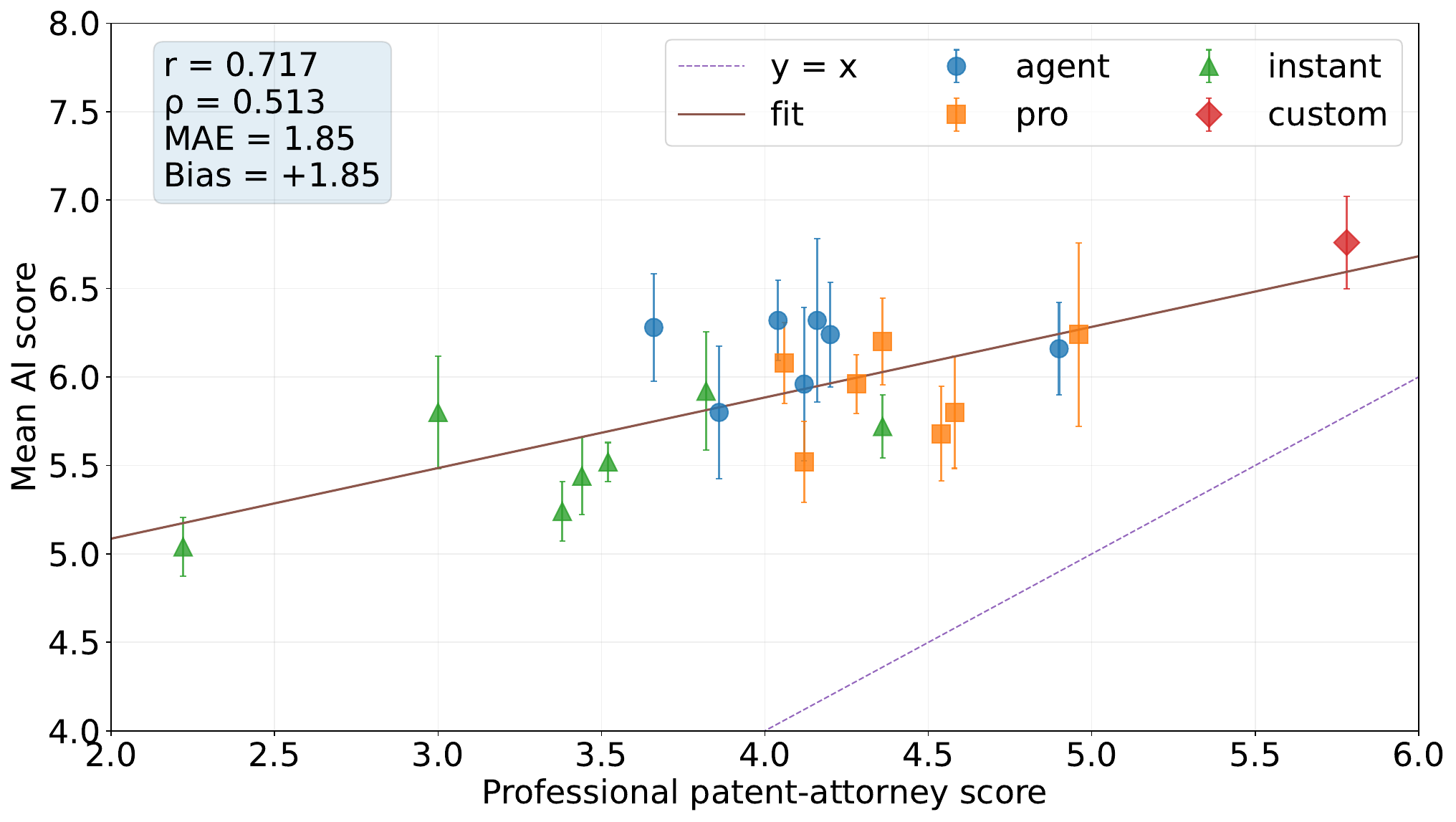}
        \vspace{-1mm}
        \textbf{(b)} LLM judge vs.\ patent attorney
    \end{minipage}
    \caption{
    LLM judge as an evaluator and optimization signal.
    \textbf{(a)} Overall judge score over revision rounds for
    Instant and Extra-High reasoning with and without structured QA feedback,
    averaged across nine inventions.
    \textbf{(b)} LLM-QA score vs.\ independent professional
    patent-attorney evaluation.
    }
    \label{fig:judge-results}
\end{figure*}

As shown in Fig.~\ref{fig:judge-results}(a), generic revision produces
substantial initial improvements but tends to saturate, whereas judge-guided
revision continues to improve through round~4. 
For Instant reasoning, the
overall judge score increases from $5.43$ to $6.71$ with QA guidance, compared
with $6.33$ under generic revision. 
Extra-High with QA increases from $6.26$ to $7.18$, compared with $6.73$ without QA.

Notably, iterative feedback substantially closes the initial reasoning gap.
Although Instant begins $0.84$ points below Extra-High, Instant+QA reaches
$6.71$ by round~4, nearly matching Extra-High without QA ($6.73$). Thus,
structured evaluator feedback can partially trade revision depth for
single-pass inference strength. 
However, these improvements in LLM-judge scores should not by themselves be interpreted as equivalent
improvements under professional evaluation.

\subsection{Agreement with Professional Evaluation}
\label{sec:human-validation}

Finally, we test whether the LLM judge reflects professional judgment.
A professional patent attorney independently scores matched patent drafts
using the same five quality dimensions. 
For each draft, we compare the attorney score with the mean score of five
independent LLM-QA evaluations.
Figure~\ref{fig:judge-results}(b) and
Table~\ref{tab:human-correlation} show significant overall association (Pearson $r=0.717$), but
agreement is strongly metric dependent. Figure Quality exhibits the strongest
relationship, and Disclosure Strength also tracks professional judgments,
whereas several other dimensions show weak correlation.

Correlation and calibration further capture different properties of the
judge. 
Claim Strength, for example, exhibits weak correlation ($r=0.103$) but small
absolute error and little systematic bias ($0.06$). Conversely, Figure and Disclosure
scores better track relative expert judgments while being systematically
higher than the attorney scores. Thus, an LLM judge may provide a useful
ranking or optimization signal without its numerical scores being calibrated
to professional ratings.

These results support LLM-based QA as a scalable experimental evaluator, but
also expose an important limitation of judge-guided optimization: increasing
the judge's score does not establish corresponding improvement under
professional judgment. Metric-specific calibration and independent expert
validation therefore remain important for professional agentic workflows.

\begin{table}[t]
\centering
\small
\caption{Agreement between repeated AI QA scores and professional patent-attorney scores across 22 matched draft--condition pairs. Bias is AI minus human score.}
\label{tab:human-correlation}
\begin{tabular}{lrrrr}
\toprule
Metric & Pearson $r$ ($p$-value) & Spearman $\rho$ ($p$-value) & MAE & Bias \\
\midrule
Prosecution Resilience & 0.042 (0.853) & 0.082 (0.717) & 1.58 & +1.58 \\
Claim Strength         & 0.103 (0.649) & 0.014 (0.949) & 0.32 & +0.06 \\
Disclosure Strength    & 0.624 (0.002) & 0.527 (0.012) & 3.41 & +3.41 \\
Figure Quality         & \textbf{0.825} (0.000) & \textbf{0.693} (0.000) & 2.61 & +2.61 \\
Filing Readiness       & 0.207 (0.356) & 0.119 (0.597) & 1.57 & +1.57 \\
\midrule
Overall            & \textbf{0.717} (0.000) & 0.513 (0.015) & 1.85 & +1.85 \\
\bottomrule
\end{tabular}
\end{table}

\section{Conclusion}

We studied LLM-as-a-judge evaluation through \emph{Vibe Patenting}, a testbed for complex professional AI workflows. 
Judge-guided revision consistently improves judge-assessed patent quality and enables a low-reasoning agent to approach the performance of substantially more expensive high-reasoning generation. 
Nonetheless, comparison with a professional patent attorney reveals strongly metric-dependent agreement and calibration. 
These results suggest that LLM judges can provide useful evaluation and optimization signals, while improvements against the judge should not be assumed to translate directly into improvements under expert evaluation.

\bibliographystyle{plainnat}
\bibliography{related_work, llm_judge_refs}

\appendix
\section{Related Work}
\label{sec:related}

\subsection{LLM-as-a-Judge}

Large language models are increasingly used as scalable evaluators for open-ended generation, where conventional reference-based metrics often correlate poorly with human judgment. Early influential work such as MT-Bench and Chatbot Arena showed that strong LLM judges can achieve substantial agreement with human preferences, while also exposing systematic failure modes including position, verbosity, and self-enhancement biases~\cite{zheng2023judging}. G-Eval further demonstrated that rubric-based prompting with explicit evaluation steps can improve human alignment for natural-language generation evaluation~\cite{liu2023geval}. Subsequent work has developed evaluator-specific models, including Prometheus for fine-grained rubric-conditioned scoring~\cite{kim2024prometheus} and Auto-J for generative alignment judgments with natural-language critiques~\cite{li2024autoj}. These studies establish LLM judging as a practical alternative to costly human evaluation, particularly when multidimensional criteria and free-form outputs make automatic metrics inadequate.

Recent work has increasingly focused on the \emph{reliability of the judges themselves}. LLMBar constructs adversarial instruction-following comparisons that reveal substantial variation across evaluator models and prompting strategies~\cite{zeng2024llmbar}, while JudgeBench evaluates judges on difficult knowledge, reasoning, mathematics, and coding comparisons and shows that strong general-purpose models can still struggle on objectively verifiable cases~\cite{tan2024judgebench}. Other studies examine systematic evaluator bias: self-preference can favor outputs that are more familiar to the judging model~\cite{wataoka2024selfpreference}, and panels of diverse evaluators can reduce intra-model bias while improving cost efficiency relative to a single large judge~\cite{verga2024poll}. Recent surveys synthesize these developments around reliability, consistency, bias mitigation, benchmarking, and deployment~\cite{li2025generationjudgment,gu2026surveyjudge}.

Our setting differs from most prior work in two respects. First, we evaluate \emph{complete professional artifacts} rather than short-form responses or pairwise preferences: a patent draft is judged along prosecution resilience, claim strength, disclosure strength, figure quality, and filing readiness. Second, the judge is not used only for offline evaluation; its structured critique is fed back to the drafting agent as an optimization signal over multiple revision rounds. We therefore study both whether an LLM judge correlates with professional patent-attorney evaluation and how judge-guided optimization differs from generic self-revision, providing a professional testbed for the reliability of LLM judges inside iterative agentic workflows.

\paragraph{LLM agents and domain-expert workflows.}
Large language models have increasingly been extended from passive text generators to agents that reason, use tools, and execute multi-step workflows. ReAct~\cite{yao2023react} interleaves reasoning and actions, while AutoGen~\cite{wu2023autogen} provides a general framework for orchestrating conversations among multiple customizable agents. Particularly relevant to our setting, MetaGPT~\cite{hong2024metagpt} encodes human standard operating procedures into structured multi-agent workflows, demonstrating how human workflow structure can support complex LLM-based tasks. Our work follows this broader direction but studies a high-stakes professional task in which an agent must transform technical source material into a set of mutually constrained artifacts. We encode patent-domain expertise through specialized drafting and analysis procedures and a shared \emph{Problem--Insight--Solution--Effect (PISE)} representation, which organizes an invention before coordinated generation of claims, specification, and figures. Rather than treating domain expertise as an alternative to stronger models, we empirically study how agentic scaffolding interacts with model capability and inference-time reasoning effort.

\paragraph{Agentic AI for scientific discovery.}
Agentic systems are also increasingly being used to automate substantial portions of the scientific process, providing a particularly relevant precedent for professional agents operating on technical knowledge~\cite{gridach2025agenticdiscovery}. In chemistry, ChemCrow~\cite{bran2024chemcrow} augments an LLM with expert-designed chemistry tools for synthesis planning, execution, drug discovery, and materials design, while Coscientist~\cite{boiko2023coscientist} combines literature and documentation search, code execution, and laboratory automation to design and perform chemical experiments. SciAgents~\cite{ghafarollahi2025sciagents} couples multi-agent reasoning with ontological knowledge graphs to generate and refine hypotheses for materials discovery. More general research agents target longer portions of the scientific workflow: Agent Laboratory~\cite{schmidgall2025agentlab} automates literature review, experimentation, and report writing from a human-provided research idea, whereas The AI Scientist~\cite{lu2026aiscientist} spans idea generation, implementation, experimentation, analysis, manuscript writing, and automated review. Recent systems move further toward iterative scientific reasoning and validation. Co-Scientist~\cite{gottweis2026coscientist} uses specialized agents for hypothesis generation, reflection, ranking, and evolution, together with scalable test-time computation and experimental validation, while Robin~\cite{ghareeb2026robin} closes a laboratory-in-the-loop cycle by connecting literature-based hypothesis generation, experimental data analysis, and subsequent hypothesis refinement. Collectively, these systems suggest that scientific agents benefit from explicit workflow decomposition, domain-specific tools and representations, and iterative evaluation rather than monolithic generation alone. Our work is complementary: instead of automating scientific discovery itself, we study the downstream transformation of scientific and technical evidence into a coordinated professional intellectual-property artifact, using PISE as an invention-centered intermediate representation and evaluating how agentic structure interacts with model capability, inference-time reasoning, and external QA-guided revision.

\paragraph{Test-time scaling and iterative refinement.}
A complementary line of research improves LLM performance by allocating additional computation at inference time. Self-consistency~\cite{wang2023selfconsistency} samples multiple reasoning trajectories and aggregates their answers, while Tree of Thoughts~\cite{yao2023tree} explicitly searches over intermediate reasoning states. Snell et al.~\cite{snell2024scaling} systematically study test-time compute scaling and show that its effectiveness depends on problem difficulty and the strategy used to allocate inference compute. These findings motivate our evaluation across multiple reasoning-effort settings: we do not assume that domain-specific agent design replaces test-time scaling, but instead examine model capability, reasoning effort, and professional agentic structure as complementary dimensions of performance.

Iterative feedback provides another form of test-time improvement. Self-Refine~\cite{madaan2023selfrefine} repeatedly generates feedback on an LLM's own output and uses that feedback for revision, while Reflexion~\cite{shinn2023reflexion} uses linguistic feedback and reflective memory to improve subsequent agent behavior. Our QA-guided revision loop is related in spirit but separates the drafting and evaluation roles: an external patent-quality agent evaluates a complete draft along multiple professional criteria, and its structured report is subsequently provided to a drafting agent for the next revision. We evaluate this process over multiple rounds and across different drafting configurations.

\paragraph{AI for patent generation and evaluation.}
Patent-language generation has recently attracted attention as a specialized application of LLMs. Jiang et al.~\cite{jiang2025claims} systematically evaluate LLM-based patent-claim generation and find that general-purpose frontier models can generate strong first independent claims, while dependent claims remain substantially more challenging and expert revision is still required. Evaluation itself is difficult because conventional text-generation metrics do not capture the structural, technical, and legal properties of patent claims. Patent-CE and PatClaimEval~\cite{jiang2025patentce} introduce expert-annotated, multidimensional evaluation of generated claims, and PatentScore~\cite{yoo2025patentscore} develops a structured claim-evaluation framework that reports strong correlation with expert annotations. These works motivate our use of multidimensional quality assessment and expert validation. Our evaluation differs in scope by assessing complete patent drafts rather than isolated claims, including prosecution resilience, claim strength, disclosure strength, figure quality, and filing readiness, and by directly comparing repeated AI assessments with scores from a professional patent attorney.

Commercial interest in AI-assisted patent preparation has also grown rapidly. Current products include Patsnap Eureka IP~\cite{patsnap2026eureka}, DeepIP~\cite{deepip2026drafting}, Patlytics~\cite{patlytics2026drafting}, Rowan Patents~\cite{clarivate2026rowan}, and LexisNexis PatentOptimizer~\cite{lexisnexis2026patentoptimizer}. Public product descriptions span invention-disclosure processing, prior-art analysis, claim and specification drafting, drawing support, consistency checking, and prosecution assistance. These systems demonstrate substantial practical demand for AI-assisted patent workflows, but their internal architectures, model and reasoning configurations, evaluation protocols, and controlled comparisons are generally proprietary. Consequently, our objective is not to claim the first use of AI for patent drafting, but to use patent drafting as a controlled testbed for studying how \emph{model capability, inference-time reasoning, structured domain expertise, and iterative QA-guided refinement} jointly affect the quality and efficiency of professional AI agents.

\section{Vibe Patenting: Agentic Framework}
\label{app:framework}

We develop \emph{Vibe Patenting}, an agentic AI framework that transforms scientific and technical source materials into a coordinated patent draft package. Unlike zero-shot prompting, in which an LLM directly converts source text into patent prose, our framework explicitly decomposes patent drafting into invention understanding, protection-oriented reasoning, document generation, and quality assurance. The system accepts heterogeneous technical evidence---such as research papers, technical reports, slides, experimental results, and source code---and produces claims, a specification, patent figures, and supporting analysis reports. Figure~\ref{fig:vibe-framework} illustrates the overall workflow.

\subsection{From Technical Evidence to Patent Package}

Let $\mathcal{X}$ denote a collection of technical source materials describing a candidate invention. The objective is to produce a patent package
$    \mathcal{D}
    =
    \{\mathcal{C},\mathcal{S},\mathcal{F},\mathcal{A}\}$,
where $\mathcal{C}$ denotes claims, $\mathcal{S}$ the written specification, $\mathcal{F}$ patent figures, and $\mathcal{A}$ auxiliary analyses such as invention characterization, prior-art-oriented observations, embodiment expansion, support analysis, and quality assessment.

This formulation differs from ordinary long-form generation because the output components are mutually constrained. Independent claims should capture the inventive concept while maintaining support in the written description; dependent claims should provide meaningful fallback positions; embodiments should broaden implementation coverage; figures should correspond to the terminology and structures described in the specification; and all components should remain consistent with the technical evidence. We therefore treat patent generation as a coordinated reasoning problem rather than independent document generation.
The overall process consists of four stages:
\begin{enumerate}
    \item \textbf{Technical evidence ingestion:} analyze the source material and extract technical contributions, assumptions, mechanisms, experimental evidence, and candidate inventive concepts.
    \item \textbf{Invention structuring:} organize the candidate invention through the PISE representation described below.
    \item \textbf{Patent package generation:} invoke specialized patent-oriented skills for protection strategy, claim drafting, specification drafting, embodiment expansion, figure generation, and supporting analysis.
    \item \textbf{Quality assurance and revision:} evaluate the complete draft using an external QA process and, when necessary, feed the resulting analysis back to the drafting agent.
\end{enumerate}

\subsection{PISE: A Structured Invention Representation}
\label{sec:pise}

A central component of both of our agentic implementations is the \emph{Problem--Insight--Solution--Effect (PISE)} representation. Given technical evidence $\mathcal{X}$, the agent first constructs $
    \mathcal{Z}_{\mathrm{PISE}}
    =
    (P,I,S,E)$,
where:
\begin{itemize}
    \item $P$ (\textbf{Problem}) describes the technical limitation, unmet need, or deficiency being addressed;
    \item $I$ (\textbf{Insight}) captures the key technical recognition or inventive idea that enables a solution;
    \item $S$ (\textbf{Solution}) describes the technical mechanism, structure, procedure, or combination that realizes the insight; and
    \item $E$ (\textbf{Effect}) identifies the technical consequences, improvements, capabilities, or measurable advantages produced by the solution.
\end{itemize}

PISE is intended to make latent invention structure explicit before patent text is generated. In particular, the four elements are constrained semantically rather than being independent summaries. The solution should address the problem through the identified insight, and the stated effects should be technically attributable to the solution. These relations can be expressed abstractly as
$    I \Rightarrow S,
    S \models P,
    S \Rightarrow E$,
where the notation represents semantic consistency rather than formal logical implication.

A technical work may contain multiple related inventive concepts. In such cases, the system can construct a hierarchy or collection of PISE units,
$    \mathcal{Z}
    =
    \{\mathcal{Z}^{(1)}_{\mathrm{PISE}},
      \ldots,
      \mathcal{Z}^{(K)}_{\mathrm{PISE}}\}$,
allowing the agent to separate a broad inventive concept from narrower implementations and alternative embodiments. This structured representation becomes the shared reasoning state used by downstream drafting modules.

The motivation for PISE is that papers and technical reports are generally organized to communicate scientific results, whereas patents must organize the same technical knowledge around protectable inventive concepts. PISE provides an intermediate abstraction between these two document structures.

\subsection{Professional Patent-Drafting Skills}

After construction of the invention representation, the agent invokes a collection of patent-oriented reasoning and generation skills. The workflow used in our system includes  technical insight and invention analysis; prior-art-oriented analysis; strength and weakness assessment; protection and claim strategy; alternative embodiment generation;  independent and dependent claim drafting; specification drafting; patent figure generation; support and consistency analysis; and quality review and revision.

These skills are not intended to act as independent text generators. They operate on a shared representation of the invention and on artifacts produced by earlier stages. For example, claim drafting uses the PISE representation together with the selected protection strategy; specification drafting expands the same concepts into enabling embodiments; and figure generation converts important structures and relationships into graphical form consistent with the specification.

The resulting computation can be summarized as
$
    \mathcal{X}
    \xrightarrow{f_{\mathrm{inv}}}
    \mathcal{Z}_{\mathrm{PISE}}
    \xrightarrow{f_{\mathrm{strategy}}}
    \mathcal{R}
    \xrightarrow{f_{\mathrm{draft}}}
    \mathcal{D}$,
where $\mathcal{R}$ represents intermediate patent strategy and embodiment decisions. This decomposition enables the system to perform explicit invention-oriented reasoning before committing to detailed patent language.

\subsection{Professional Agent Implementations}
\label{sec:agent-architectures}

\begin{figure*}[t]
    \centering
    \includegraphics[width=\textwidth,trim=0 150 0 0,clip]{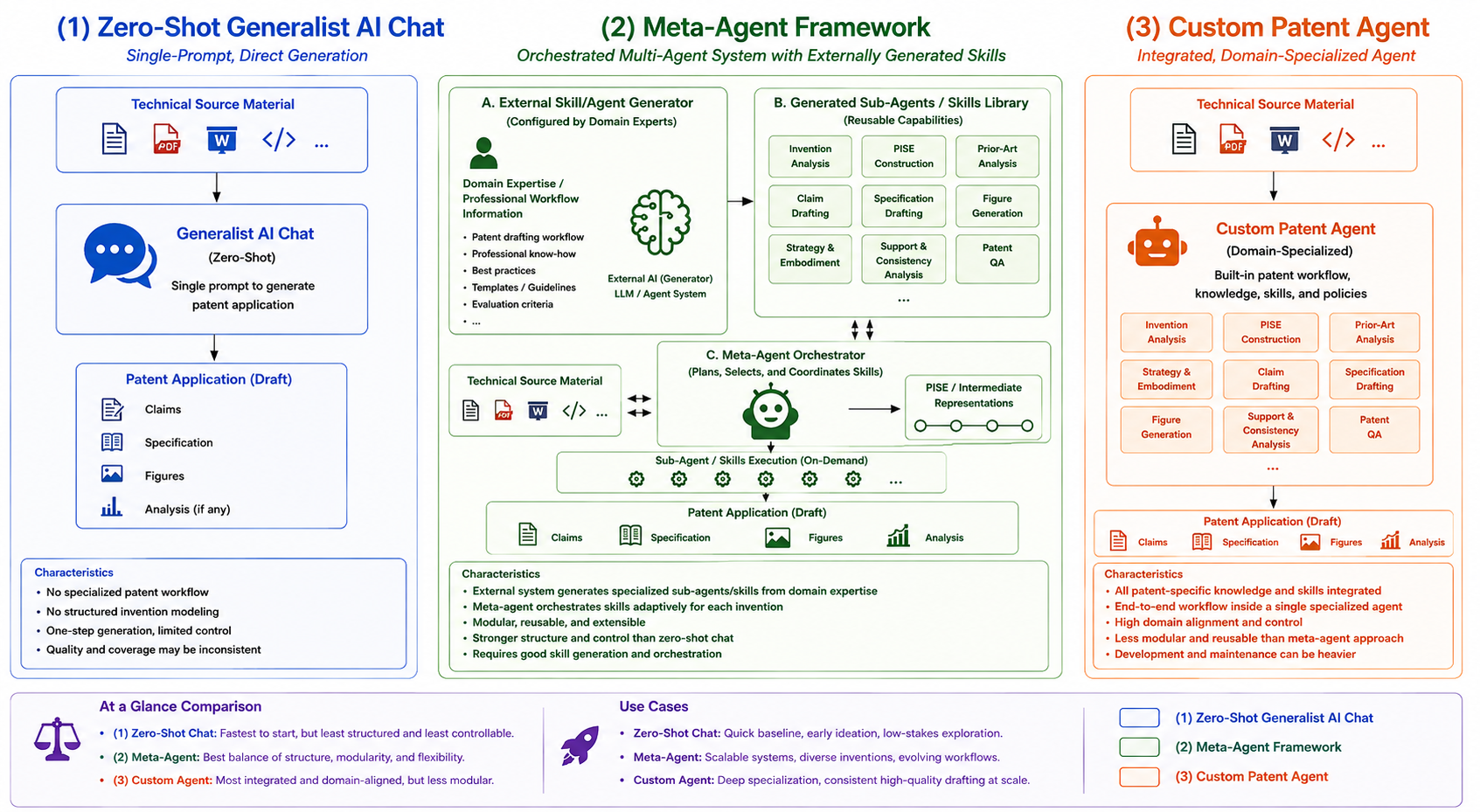}
    \caption{
    Three agent architectures. 
    The generalist AI chat reasons about how to convert technical materials into patent drafts in zero-shot fashion.
    The meta-agentic framework constructs reusable patent-oriented skills and workflow to orchestrate. 
    The custom patent agent directly integrates patent-specific workflow logic, and specialized drafting skills. 
    }
    \label{fig:architectures}
\end{figure*}

We investigate three implementations of professional AI patent drafting, illustrated in Figure~\ref{fig:architectures}.

\paragraph{Generalist chat AI.}
The simplest approach uses a modern general-purpose chat LLM to convert technical materials directly into a patent draft in a zero-shot manner.
The AI makes its own reasoning effort to generate a patent package.

\paragraph{Meta-agentic framework.}
The second implementation is to use a general-purpose agent to generate professional skills and to orchestrate sub-agents, from expert knowledge and patent-specific workflow. 
The agent receives a high-level patent-generation task and dynamically invokes reusable skills for invention analysis, PISE construction, claim drafting, specification drafting, figure generation, and related subtasks. 
Intermediate artifacts provide context for later stages of the workflow autonomously.

\paragraph{Custom patent agent.}
The third implementation packages professional patent-drafting behavior into a dedicated domain agent. Patent-specific workflow instructions, drafting guidelines, templates, reference materials, and specialized skills are directly available to the agent. The agent is instructed to perform invention analysis, construct the PISE representation, develop a protection strategy, and coordinate the generation of the complete patent package.
This architecture places substantial domain knowledge inside the agent configuration. It therefore provides a strong form of professional scaffolding and serves as our most specialized system.

Conceptually, the custom agent encodes professional behavior primarily through a dedicated agent configuration, whereas the meta-agentic system externalizes more of this behavior into modular skills and workflow components through the use of generalist AI agent. 
This distinction allows us to examine whether professional capability can be obtained through reusable agentic scaffolding rather than only through a strongly customized domain agent.

\subsection{End-to-End Workflow}

Given technical source material $\mathcal{X}$, both of our agentic implementations execute a workflow of the general form
\begin{equation}
    \mathcal{X}
    \rightarrow
    \mathcal{Z}_{\mathrm{PISE}}
    \rightarrow
    \mathcal{R}
    \rightarrow
    \{\mathcal{C},\mathcal{S},\mathcal{F},\mathcal{A}\}
    \rightarrow
    \mathcal{Q},
\end{equation}
where $\mathcal{Z}_{\mathrm{PISE}}$ is the structured invention representation, $\mathcal{R}$ denotes protection-strategy and embodiment decisions, $\mathcal{C}$ denotes claims, $\mathcal{S}$ the specification, $\mathcal{F}$ the figures, $\mathcal{A}$ supporting analyses, and $\mathcal{Q}$ the QA report.

Table~\ref{tab:skills} summarizes the major skills used in the system.

\begin{table*}[t]
\centering
\small
\caption{Representative professional skills used by the Vibe Patenting framework.}
\label{tab:skills}
\begin{tabular}{p{0.19\textwidth}p{0.29\textwidth}p{0.22\textwidth}p{0.22\textwidth}}
\toprule
\textbf{Skill} & \textbf{Purpose} & \textbf{Primary Input} & \textbf{Output} \\
\midrule
Technical analysis
& Identify contributions, mechanisms, assumptions, and evidence
& Paper/report/slides/code
& Structured technical findings
\\

Invention discovery
& Identify potentially patentable concepts and abstractions
& Technical findings
& Candidate inventions
\\

PISE construction
& Represent each invention as Problem--Insight--Solution--Effect
& Candidate invention + evidence
& PISE representation
\\

Prior-art-oriented analysis
& Identify likely novelty and obviousness pressure points
& PISE + technical context
& Prior-art-oriented analysis
\\

Strength/weakness analysis
& Identify protection opportunities, gaps, and design-around risks
& PISE + analysis
& Strategic assessment
\\

Patent strategy
& Determine protection hierarchy and independent-claim concepts
& PISE + strategic assessment
& Claim/protection strategy
\\

Embodiment expansion
& Generate alternative implementations and fallback positions
& PISE + technical context
& Embodiment set
\\

Claim drafting
& Generate independent and dependent claim hierarchy
& PISE + strategy + embodiments
& Claims
\\

Specification drafting
& Generate enabling disclosure consistent with claims
& Claims + PISE + embodiments
& Specification
\\

Figure generation
& Generate patent-oriented system, process, and embodiment diagrams
& PISE + claims + specification
& Patent figures
\\

Support/consistency analysis
& Check terminology, claim support, and figure/spec alignment
& Complete draft
& Support analysis
\\

Patent QA
& Score and critique complete patent package
& Draft + figures
& Multi-axis scores + report
\\

Revision
& Address identified QA weaknesses
& Draft + QA report
& Revised draft
\\
\bottomrule
\end{tabular}
\end{table*}

\subsection{Custom Patent Agent}

The custom implementation integrates the patent-specific workflow, professional instructions, reference materials, templates, and specialized skills into a dedicated domain agent. The agent is explicitly instructed to reason about the technical material before drafting and to generate a coordinated patent package rather than independent document fragments.

The custom agent therefore has access to:
\begin{itemize}
    \item patent-drafting workflow instructions;
    \item structured PISE reasoning;
    \item patent strategy and embodiment-expansion procedures;
    \item claim and specification drafting skills;
    \item patent-figure generation procedures;
    \item support and quality-analysis skills; and
    \item revision procedures for addressing QA feedback.
\end{itemize}

The custom implementation represents a strongly domain-specialized configuration in which professional knowledge is embedded directly into the agent environment.

\subsection{Meta-Agentic Implementation}

The meta-agentic implementation separates the general orchestration agent from the patent-specific skills. 
The LLM agent first constructs professional skills and workflow logic from expert information.
A general-purpose agent then dynamically invokes reusable capabilities/skills for invention analysis, PISE construction, claim generation, specification drafting, figures, and QA.

This design externalizes more of the professional knowledge into reusable skills and intermediate artifacts:
\begin{equation}
    \text{Meta-Agent}
    +
    \{\text{Patent Skills}\}
    +
    \mathcal{Z}_{\mathrm{PISE}}
    \rightarrow
    \text{Patent Package}.
\end{equation}

The distinction between the two implementations is therefore not whether professional structure is used---both systems use it---but where that structure is represented. The custom agent encodes more professional behavior in a dedicated agent configuration, whereas the meta-agentic system composes modular skills through a more general orchestrator.

\section{PISE Representation and Examples}
\label{app:pise}

\subsection{PISE Structure}

PISE represents an invention using
\begin{equation}
    \mathcal{Z}_{\mathrm{PISE}}
    =
    (P,I,S,E),
\end{equation}
where $P$, $I$, $S$, and $E$ denote Problem, Insight, Solution, and Effect, respectively.

\begin{figure}[t]
    \centering
    \includegraphics[width=\columnwidth,trim=0 60 0 150,clip]{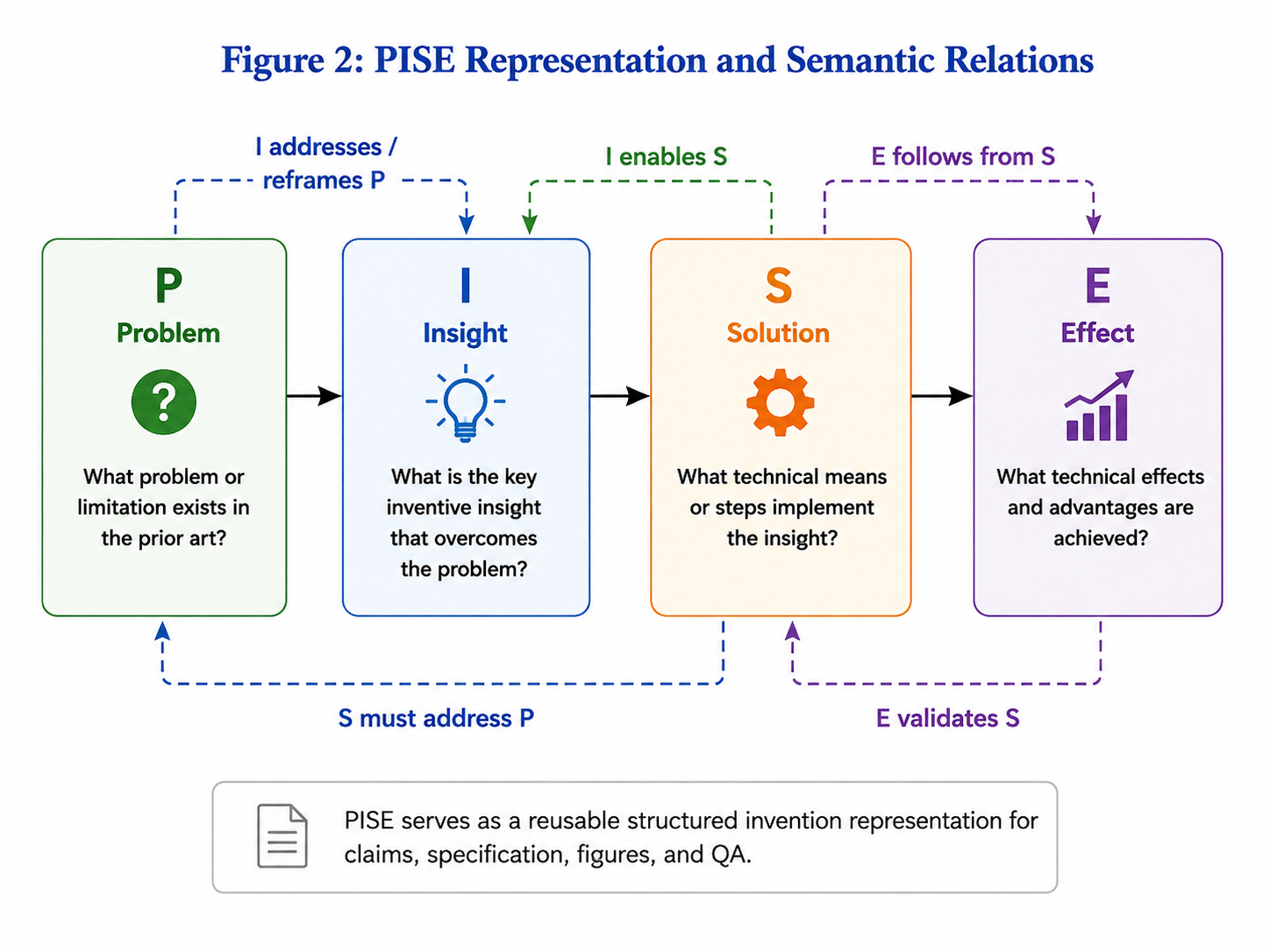}
    \caption{
    PISE structured invention representation. The solution should address the identified problem through the inventive insight, while the resulting technical effect should follow from the solution. The same representation is reused for patent strategy, claims, specification, figures, and QA.
    }
    \label{fig:pise}
\end{figure}

The purpose of PISE is not simply to summarize a technical paper. Instead, it reorganizes the source material according to invention-oriented semantics:
\begin{itemize}
    \item the \textbf{Problem} identifies the relevant technical limitation;
    \item the \textbf{Insight} captures the inventive recognition that enables a new approach;
    \item the \textbf{Solution} identifies the technical mechanism implementing that insight; and
    \item the \textbf{Effect} describes the resulting technical consequence or improvement.
\end{itemize}
Figure~\ref{fig:pise} shows the PISE representation and its semantic relations.
Specifically, the AI agent makes a reasoning process: what problem/limitation exists in the prior art; what is the key invention insight that overcomes the problem; what technical steps/methods implement the insight; what technical effects/advantages are achieved.

The following simplified example illustrates the representation format. The text is schematic and is intended to show structure rather than reproduce any confidential invention.

\begin{quote}
\textbf{Problem:}
Existing inference-time compression methods may rely on fixed transformations that do not adapt to the structure of the particular input or activation instance.

\textbf{Insight:}
The compressibility of a tensor can depend strongly on how its dimensions or indexed components are organized, and this organization can be adapted using information available at inference time.

\textbf{Solution:}
Determine an input-dependent ordering or transformation of tensor components, apply the resulting transformation, and compress the transformed representation using a structured low-complexity model.

\textbf{Effect:}
The transformed representation can exhibit lower effective complexity, enabling reduced storage or computation while preserving inference quality.
\end{quote}

The structured invention model can then be reused across downstream artifacts:
\begin{equation}
    \mathcal{Z}_{\mathrm{PISE}}
    \rightarrow
    \begin{cases}
        \text{claim scope},\\
        \text{fallback limitations},\\
        \text{embodiments},\\
        \text{specification},\\
        \text{figures},\\
        \text{QA criteria}.
    \end{cases}
\end{equation}

\subsection{Multiple PISE Units}

A technical work may support multiple inventive concepts. In such cases, the system maintains a set
\begin{equation}
    \mathcal{Z}
    =
    \left\{
    \mathcal{Z}^{(1)}_{\mathrm{PISE}},
    \ldots,
    \mathcal{Z}^{(K)}_{\mathrm{PISE}}
    \right\}.
\end{equation}

These units may represent a broad parent invention, narrower implementation mechanisms, alternative embodiments, hardware-specific realizations, or complementary inventions. The drafting workflow can selectively combine these units when constructing independent claims and fallback positions.

\begin{figure}
    \centering
    \includegraphics[width=\linewidth]{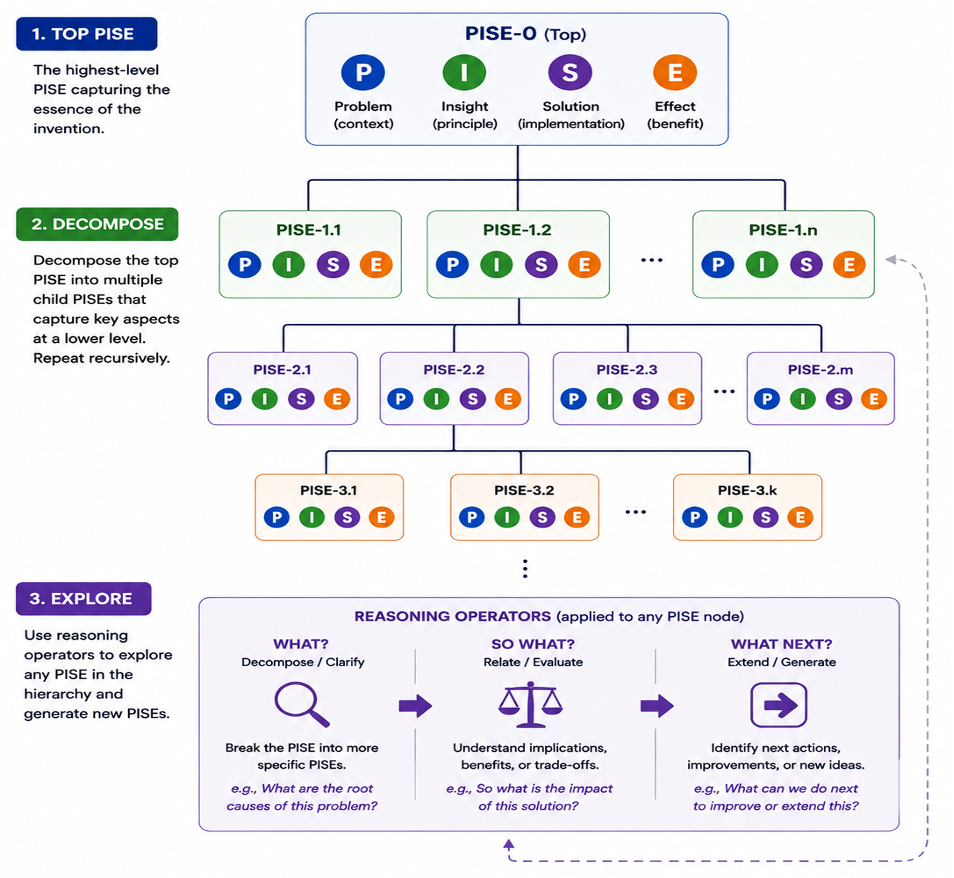}
    \caption{PISE hierarchy for constrained exploration of unconstrained knowledge:
    Extract top PISE core; extend PISE trees; and explore new PISE nodes through reasoning. }
    \label{fig:pise_tree}
\end{figure}

Figure~\ref{fig:pise_tree} illustrates the PISE hierarchy. 
From the technical evidence/materials, the agent first extracts a set of PISE units, and extends them into one core unified top PISE to cover all PISE units. 
The agent also explores new PISE units through reasoning: what are the root causes of problems; what is the impact of the solution; how the solution can be extended or improved.

\section{AI Judge for Patent Quality Assurance (QA)}
\label{app:judge}

\subsection{Patent QA Skill}

We built a reusable patent QA skill, which performs rigorous pre-filing quality assurance of U.S.\ utility patent drafts under USPTO practice.
It produces an attorney-grade report with strict integer-only 1-10 scores for five defined axes and supports hybrid draft-based plus web-assisted review.
It attempts to identify defects that could reduce claim value, create prosecution difficulty, weaken \S112 support, impair enforceability, or make the package unready to file.
It always analyzes the actual draft and figures before scoring without inferring quality from polish alone.

Review workflow follows the sequence:
\begin{enumerate}
\item Inventory the supplied files and identify the primary patent draft, claim set, complete patent figure set, and any supplemental technical sources.
\item Read the entire patent draft, including claims, specification, abstract, tables, comments, tracked changes, placeholders, and embedded figures when accessible.
\item Inspect every patent figure visually at useful resolution. Do not rely only on extracted text or alt text.
\item If a supplemental technical report, invention disclosure, experimental report, or academic-paper draft is supplied, read it carefully as a separate technical source, including its figures, tables, equations, appendices, and experimental details. Read references/technical-source-crosscheck.md.
\item Build a working model of the invention using all supplied technical information, while preserving a strict distinction between what the patent application itself discloses and what appears only in supplemental material.
\item Cross-check claims against the patent specification and patent figures. For each material independent-claim limitation, locate written-description support, enabling disclosure, and relevant patent-figure support where appropriate. Do not count report-only material as patent-application support.
\item Cross-check the patent draft against supplemental technical sources for omitted embodiments, mechanisms, variants, parameters, terminology, experimental evidence, technical advantages, and inconsistencies that matter to claim scope or filing value.
\item Analyze prosecution and claim risks under current U.S. utility patent practice. Read legal-research-framework.md when evaluating \S101/102/103/112, prior art, legal guidance, web research, or possible publication/public-disclosure issues.
\item Perform or recommend web-assisted review as described below. Keep intrinsic patent-draft findings, supplemental-source findings, and web-assisted findings distinct.
\item Review the patent figure set for substantive coverage, clarity, consistency, reference numerals, and filing-quality issues.
\item Review filing-readiness defects: unresolved drafting notes, inconsistent terminology, broken dependencies or antecedent basis, missing sections/figures, mismatched numbering, unsupported statements, report-only technical content that should be incorporated before filing, and other substantive or formal issues visible from the supplied package.
\item Prioritize findings by severity, then assign scores only after the qualitative analysis is complete. Read scoring-rubric.md before scoring.
\item Create the final report using report-format.md.
\end{enumerate}

The review standards are as follows:

\paragraph{Claims}
Review every independent claim closely and the dependent-claim strategy as a set. Evaluate at least:
\begin{itemize}
    \item 
scope relative to the disclosed inventive contribution;
\item unnecessary narrowing and accidental overbreadth;
\item functional/result-oriented language and structural or procedural support;
\item antecedent basis, dependency, clarity, and internal claim consistency;
\item claim categories and coverage of commercially relevant implementations;
\item fallback positions and useful dependent limitations;
\item enforceability concerns, divided-infringement or actor issues when relevant;
\item design-around opportunities and whether the claims protect the actual inventive center;
\item support for each material limitation in the specification and figures;
\item terminology drift between claims, specification, and figures.
\end{itemize}
Do not reward breadth by itself. Broad claims unsupported by disclosure or exposed to known art are weak claims.

\paragraph{Disclosure}
Evaluate \S112 support from the patent application itself across the intended claim scope. Supplemental technical sources may reveal what is missing, but they do not cure an omission unless the substance is included in the application before filing. Evaluate at least:
\begin{itemize}
\item written description for claimed combinations and alternatives;
\item enablement across the claimed scope without undue experimentation;
\item implementation detail proportionate to predictability of the technology;
\item embodiments, alternatives, ranges, optionality, substitutions, and fallback disclosure;
\item definitions and consistent use of important terms;
\item support for functional language and any means-plus-function concerns;
\item consistency among summary, detailed description, claims, abstract, and figures;
\item whether likely amendments during prosecution would have clear original support.
\end{itemize}
Distinguish missing disclosure that cannot safely be added after filing from ordinary editorial improvements. Treat the former as much more serious.

\paragraph{Figures}
Evaluate both substantive coverage and drawing quality. Check whether the figures:

\begin{itemize}
\item cover each important embodiment, architecture, flow, state, component relationship, or variation that materially supports the claims;
\item use consistent FIG. numbers, reference numerals, labels, arrows, and terminology;
\item correspond to the written description and brief description of the drawings;
\item avoid unexplained elements, orphan numerals, missing numerals, and conflicting labels;
\item are legible and understandable without guessing;
\item provide enough visual disclosure to support key structural/functional relationships;
\item appear suitable for filing or need formal drawing cleanup.
\end{itemize}
Do not equate visual neatness with substantive figure completeness.

\subsection{Scoring Rubric}

Score calibration is as follows:
\begin{itemize}
\item[Score]	General meaning
\item[10]	Exceptional and essentially filing-ready on this axis; no material deficiency found. Reserve for rare drafts.
\item[9]	Filing-ready on this axis with only minor, non-substantive cleanup.
\item[8]	Strong; limited meaningful improvements remain, but no major weakness.
\item[7]	Generally solid but meaningful revisions are advisable before filing.
\item[6]	Material weaknesses exist and should be addressed before filing.
\item[5]	Multiple material weaknesses or one major weakness; not comfortably filing-ready.
\item[4]	Significant deficiencies affect value, support, prosecution, or completeness.
\item[3]	Major systemic deficiencies create substantial prosecution or validity risk.
\item[2]	Severe deficiencies across core aspects of the axis.
\item[1]	Fundamentally deficient; substantial reconstruction is required.
\end{itemize}

\subsection{Patent Quality Dimensions for QA Scoring}
\label{app:quality-dimensions}

Our QA framework evaluates a patent draft along five dimensions:
\begin{enumerate}
    \item \textbf{Prosecution Resilience}: withstanding prosecution challenges, including prior art, novelty, obviousness, eligibility, and other  \S101/102/103 risks;
    \item \textbf{Claim Strength}: quality and breadth of the claims, including independent-claim scope, fallback positions, enforceability, and design-around resistance;
    \item \textbf{Disclosure Strength}: strength of \S112 support, including written description, enablement, embodiments, alternatives, terminology, and internal consistency;
    \item \textbf{Figure Quality}: 
    completeness and clarity of the figure set, including coverage of key embodiments/variations, readability, consistency, and support for the specification and claims; and
    \item \textbf{Filing Readiness}: overall readiness for filing, considering substantive gaps, drafting issues, claims, specification, and figures.
\end{enumerate}

Each dimension is scored on a ten-point scale. 
The meta-agent built an independent agent skill for QA evaluations, and we use mean score among 5 QA agents. 
This multidimensional evaluation enables us to compare general-purpose chat models, reasoning-effort settings, agentic systems, model generations, and QA-guided revision under a common quality framework. We additionally compare the AI assessments with evaluations from a professional patent attorney.

\section{QA Judge-Guided Revision}
\label{app:qa-loop}

Generating a complete patent package in a single pass can leave weaknesses such as insufficient support, overly narrow or broad claims, terminology inconsistencies, missing embodiments, or inadequate figure coverage. We therefore introduce a QA-guided revision loop in which the drafting and evaluation functions are separated.

Let $\mathcal{D}_{t}$ denote the patent draft produced at revision round $t$. An independent QA agent evaluates the draft and produces a structured report $
    \mathcal{Q}_{t}
    =
    f_{\mathrm{QA}}(\mathcal{D}_{t})$,
where $\mathcal{Q}_{t}$ contains numerical assessments together with identified weaknesses and revision suggestions.

The drafting agent then receives the current patent draft, the corresponding QA report, and the original technical context:
$    \mathcal{D}_{t+1}
    =
    f_{\mathrm{rev}}
    \left(
        \mathcal{X},        \mathcal{Z}_{\mathrm{PISE}},
        \mathcal{D}_{t},
        \mathcal{Q}_{t}
    \right)$.
The process can be iterated for several rounds:
$
    \mathcal{D}_{1}
    \rightarrow
    \mathcal{Q}_{1}
    \rightarrow
    \mathcal{D}_{2}
    \rightarrow
    \mathcal{Q}_{2}
    \rightarrow \cdots
    \rightarrow
    \mathcal{D}_{T}$.

\begin{figure}[t]
    \centering
    \includegraphics[width=\columnwidth]{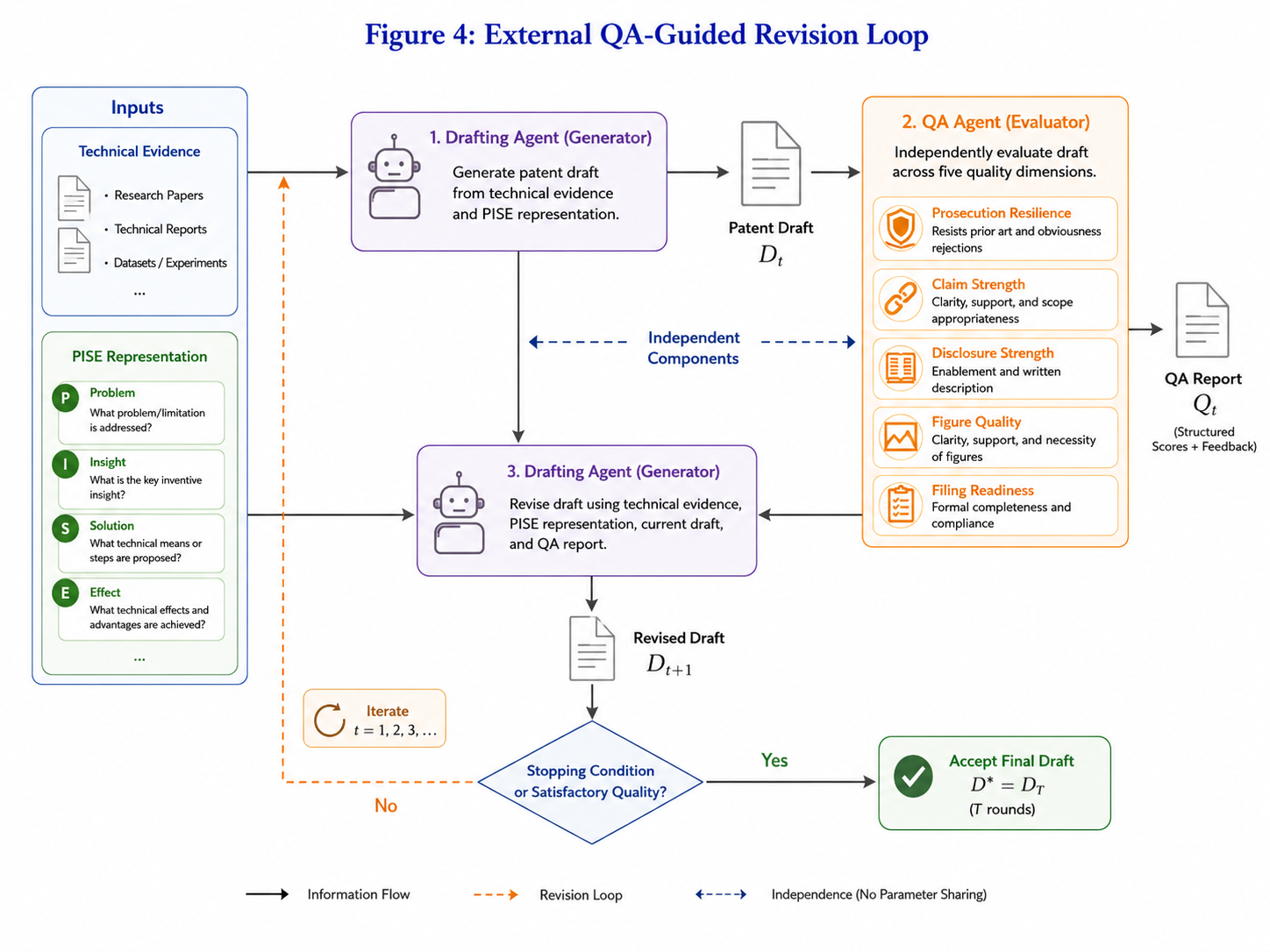}
    \caption{
    External QA-guided revision loop. A drafting agent first generates a patent draft from the technical evidence and structured invention representation. A separate QA process evaluates the complete draft and returns structured scores and feedback. The drafting agent then revises the patent using the QA report, and the process can be repeated for multiple rounds.
    }
    \label{fig:qa-loop}
\end{figure}

Figure~\ref{fig:qa-loop} illustrates this external revision mechanism.

This setup differs from simply prompting the drafting agent to ``improve'' its own output. The QA report provides a stable domain-specific interface between evaluation and generation, allowing the same evaluation procedure to guide different models and agent configurations. It also enables us to experimentally measure how patent quality evolves as additional revision rounds are allocated.

\subsection{Agent Instructions and Prompting}
\label{app:prompts}

This section documents the principal prompting interfaces used in our experiments. We distinguish between the minimal baseline prompts used for generic LLM generation and the richer instructions available to the professional agentic systems.

\subsubsection{Instruction Prompting}

Each agent receives the source technical material together with a concise request to generate a patent draft. The prompt intentionally does not expose the full professional workflow or modular patent skills used by the proposed systems.
All agents receive the same instruction prompts as follows:
\begin{quote}
\small
\texttt{Use this technical report to create a full patent draft package, including prior-art-oriented analysis, strengths and weaknesses, embodiment expansion, 3 independent claims, 17 dependent claims, specification sections, a patent draft markdown file, a patent draft DOCX file, direct high-resolution black-and-white patent figure images, the corresponding figure generation prompt markdown file, and a patent-analysis note.}
\end{quote}

The same task is used across reasoning-effort settings whenever possible so that the principal changed variable is the agent configuration, the model or inference configuration.

\subsubsection{Patent QA Prompt}

The QA agent receives a completed patent draft and, when available, associated patent figures. It evaluates five dimensions on a ten-point scale and provides structured justification and recommendations.
The instruction prompt is just to invoke the patent-qa skill.

\begin{quote}
\small
\texttt{Use @patent-qa skill}
\end{quote}

\subsubsection{QA Judge-Guided Revision Prompt}

At revision round $t+1$, the drafting agent receives the previous draft $\mathcal{D}_t$ together with the QA report $\mathcal{Q}_t$.
The same instruction prompt with an additional sentence to use the QA report is given as follows:
\begin{quote}
\small
\texttt{Use this technical report to create a full patent draft package, including prior-art-oriented analysis, strengths and weaknesses, embodiment expansion, 3 independent claims, 17 dependent claims, specification sections, a patent draft markdown file, a patent draft DOCX file, direct high-resolution black-and-white patent figure images, the corresponding figure generation prompt markdown file, and a patent-analysis note. NOTE: please revise and improve the initial draft package attached according to the quality analysis report given.}
\end{quote}

We intentionally separate the QA report from the drafting process so that the same feedback interface can be used across different model and agent configurations.

\subsubsection{QA Judge-Unguided Generic Revision Prompt}

At revision round $t+1$, the drafting agent receives the previous draft $\mathcal{D}_t$ without using the QA report $\mathcal{Q}_t$.
The same instruction prompt except for the last few words is given as follows:
\begin{quote}
\small
\texttt{Use this technical report to create a full patent draft package, including prior-art-oriented analysis, strengths and weaknesses, embodiment expansion, 3 independent claims, 17 dependent claims, specification sections, a patent draft markdown file, a patent draft DOCX file, direct high-resolution black-and-white patent figure images, the corresponding figure generation prompt markdown file, and a patent-analysis note. NOTE: please revise and improve the initial draft package attached.}
\end{quote}

We intentionally separate the QA report from the drafting process so that the same feedback interface can be used across different model and agent configurations.

\section{Examples of Generated Patent Artifacts}
\label{app:artifacts}

This section provides representative outputs generated by the agentic workflow. The purpose is not to evaluate the legal merit of an individual patent application, but to illustrate the diversity and coordination of artifacts produced from a single technical input.

\subsection{Representative Patent Figures}

The framework generates patent-oriented figures rather than directly copying figures from the source publication. Depending on the invention, generated figures may include:
\begin{itemize}
    \item overall system architectures;
    \item algorithmic process diagrams;
    \item component-level architectures;
    \item training or inference workflows;
    \item hardware embodiments;
    \item alternative implementations; and
    \item interaction diagrams among system components.
\end{itemize}

\begin{figure*}[t]
    \centering
    \includegraphics[width=\linewidth]{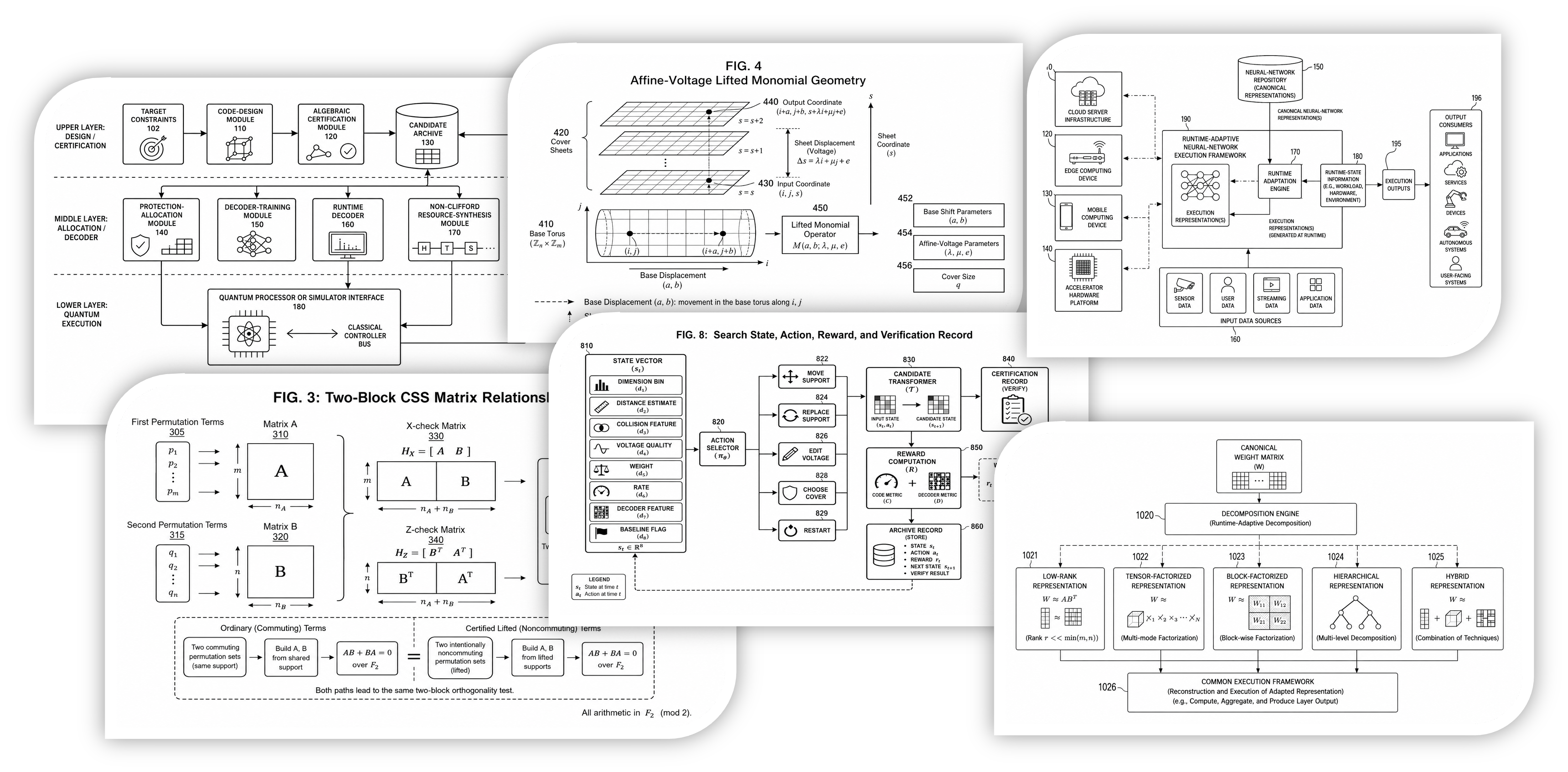}
    \caption{
    Representative patent figures generated from technical source material. The system can generate figures that reorganize or expand the technical content into patent-oriented system, process, and embodiment views.
    }
    \label{fig:artifact-example-1}
\end{figure*}

Figure~\ref{fig:artifact-example-1} shows representative outputs.

\subsection{Representative Claim Generation}

A generated patent package includes a hierarchy of independent and dependent claims. The agent first determines broad protection concepts from the PISE representation and then introduces narrower limitations as fallback positions.

\begin{quote}
\small
\textbf{Illustrative independent claim excerpt:}

\texttt{14. A non-transitory computer-readable medium storing instructions that, when executed by one or more processors, cause the one or more processors to perform operations comprising:
enumerating candidate tensor-encoding configurations, each candidate tensor-encoding configuration specifying a folding of a source tensor, a value mapping, a reduction across one or more slice modes, a permutation scope, a tensor-decomposition topology, one or more tensor ranks, and a permutation encoding;
for each of a plurality of the candidate tensor-encoding configurations:
generating a score field by applying the value mapping to values of the source tensor and applying the reduction across the one or more slice modes;
deriving, from the score field, a candidate reversible permutation within the permutation scope;
applying the candidate reversible permutation to multiple slices of the source tensor to obtain a candidate ordered tensor;
forming candidate tensor cores by decomposing the candidate ordered tensor according to the tensor-decomposition topology and the one or more tensor ranks;
determining a reconstruction-error measure and an encoded-storage measure that includes storage for the candidate tensor cores and storage for the candidate reversible permutation under the permutation encoding;
selecting a selected tensor-encoding configuration from the plurality of the candidate tensor-encoding configurations based on the reconstruction-error measures and the encoded-storage measures;
producing a compressed representation comprising selected tensor cores, selected permutation data, and a descriptor of the selected tensor-encoding configuration; and
reconstructing at least a portion of the source tensor, or performing a consumer computation corresponding to the at least a portion, using the selected tensor cores, the selected permutation data, and the descriptor.
}

\vspace{1mm}
\textbf{Illustrative dependent claim excerpt:}

\texttt{16.	The non-transitory computer-readable medium of claim 14, wherein at least one candidate tensor-encoding configuration specifies a group sorting operation that leaves a within-group order unchanged or a sequential-axis sorting operation that sorts different subsets of tensor modes in sequence.}
\end{quote}

The claim-generation skill is coordinated with specification drafting so that important claim elements are supported by corresponding embodiments and terminology.

\subsection{Representative Specification Expansion}

The specification-generation stage expands the core invention beyond the narrow implementation appearing in the technical source. Typical expansions include:
\begin{itemize}
    \item alternative architectures;
    \item alternative parameterizations;
    \item software and hardware embodiments;
    \item centralized and distributed implementations;
    \item optional processing stages;
    \item alternative ordering of operations; and
    \item variations addressing foreseeable design-arounds.
\end{itemize}

\section{Patent Analysis and QA Report Examples}
\label{app:reports}

In addition to the patent draft itself, the system produces analysis artifacts intended to expose the reasoning behind the generated protection strategy.

\subsection{Patent Analysis}

The AI agent produces a patent analysis report during patent drafting to improve the quality inside an internal review loop.
A representative analysis report contains sections such as:
\begin{itemize}
    \item Invention analysis (problem/solution/concepts/effect/evidence/missing facts);
    \item Prior-art analysis (pressure point/weakness/eligibility/entablement);
    \item Strengthening and distinction strategy (tree architecture/fallback positions/claim support crosswalk);
    \item Embodiment expansion lists;
    \item Claim support analysis (coverage/limitation/support/concern);
    \item Figure and description alignment review;
    \item Summary and recommended plan.
\end{itemize}

\begin{figure}
    \centering
    \includegraphics[width=\linewidth]{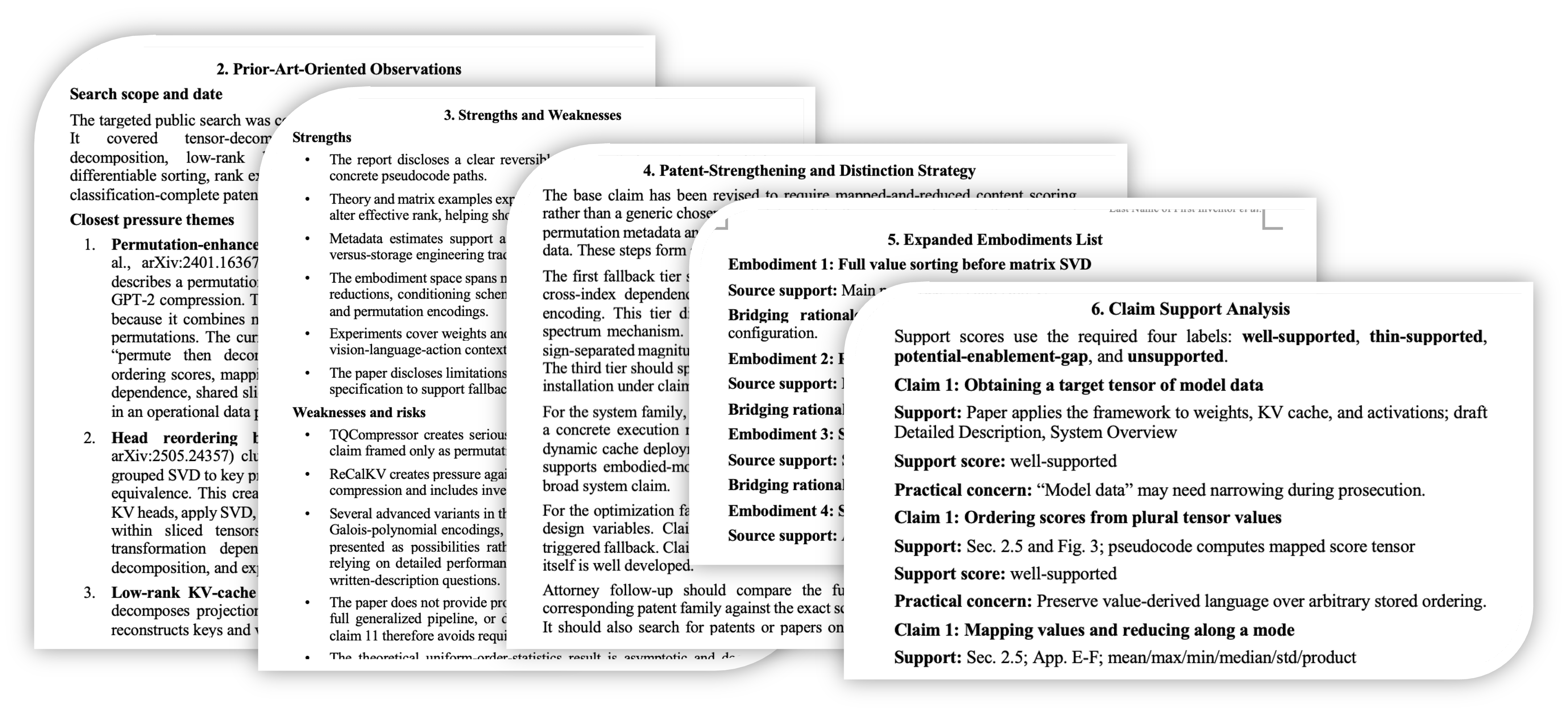}
    \caption{Sample of patent analysis report for patent drafting.}
    \label{fig:analysis}
\end{figure}

\subsection{QA Report}

The QA report is generated by a separate independent agent: LLM judge.
A representative analysis report contains sections such as:
\begin{itemize}
    \item Executive assessment;
    \item Priority issues before filing;
    \item Claim analysis (tree architecture/fallback positions/claim support crosswalk);
    \item Disclosure and \S112 analysis (description/enablement/definiteness/terminologies);
    \item Figure analysis (coverage/quality/consistency);
    \item Web-assisted research findings;
    \item Recommended revision plan;
    \item Scoring and summary.
\end{itemize}

The QA agent evaluates the complete draft along the five dimensions used in the experiments. Table~\ref{tab:qa-example} illustrates a representative scoring report.

\begin{table*}[t]
\centering
\small
\caption{Example of QA report scoring.}
\label{tab:qa-example}
\begin{tabular}{p{0.21\textwidth}c p{0.64\textwidth}}
\toprule
\textbf{Dimension} & \textbf{Score} & \textbf{Rationals}  \\
\midrule
Prosecution Resilience
& 6
& The draft has plausible technical distinctions, but the art landscape contains variational state preparation, hardware-efficient brickwork/alternating ansätze, and resource-aware compilation. A targeted screen was performed, but not an exhaustive search; obviousness risk remains material.
\\

Claim Strength
& 6
& Claim 1 captures a concrete architectural distinction, but the three independent claims protect different centers and the dependent claims need stronger combination ladders. Claim 18 carries a support/provenance concern.
\\

Disclosure Strength
& 6
& Core fixed-topology embodiments are well described and broadly supported. Drafting-origin admissions for adaptive growth and learned prediction, plus limited quantitative detail for the key empirical effects, are material weaknesses.
\\

Figure Quality
& 8
& All 20 figures were present, legible, and substantially consistent with the specification. Improvements are mainly evidentiary/formal: schematic empirical plots, unexplained FIG. 18 numbers, and a few dense drawings.
\\

Filing Readiness
& 5
& The package should not be filed unchanged because inventor-possession/provenance issues and placeholders remain, and claim strategy should be tightened. The core invention is captured, so reconstruction is not required.
\\
\bottomrule
\end{tabular}
\end{table*}

\section{Extended Experimental Results}
\label{app:experiments}

\subsection{Model, Reasoning, and Agent Scaling}

\begin{figure*}[t]
    \centering
    \includegraphics[width=\textwidth]
    {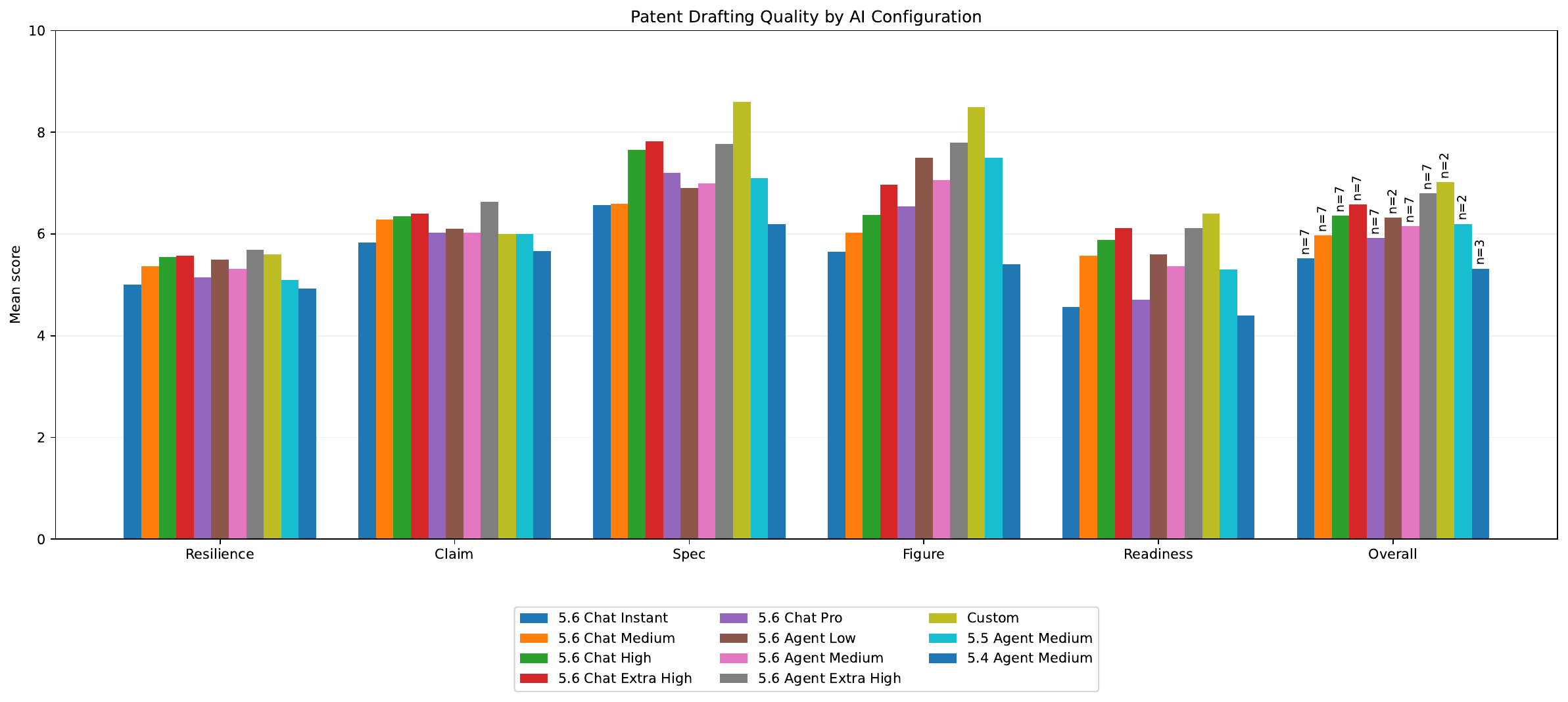}
    \caption{
    Patent quality across model, reasoning-effort, and agentic configurations. Scores are first averaged over five independent QA evaluations and then averaged over the available patent drafts. The sixth group shows the overall mean across the five quality dimensions. The number of available patents differs for several configurations and is reported with the corresponding results.
    }
    \label{fig:config-quality}
\end{figure*}

Figure~\ref{fig:config-quality} compares the five quality dimensions and overall score for the evaluated configurations.
For GPT-5.6 Sol chat mode, increasing reasoning effort generally improves drafting quality. The overall score increases from $5.53$ for Instant to $5.97$ for Medium, $6.36$ for High, and $6.58$ for Extra High. This trend supports inference-time scaling for the professional drafting task, although performance is not strictly monotonic across all available configurations; in particular, the observed Pro score is $5.93$.

Agentic execution also produces strong results. The GPT-5.6 agent obtains an overall score of $6.15$, while its Extra-High reasoning configuration reaches $6.80$. 
The GPT-5.4 and GPT-5.5 agent configurations obtain overall scores of $5.32$ and $6.20$, respectively, providing additional evidence that underlying model capability contributes substantially to professional-task performance.

The custom patent agent obtains an overall score of 7.02 across the two available patents, with particularly strong Disclosure and Figure scores. Because the custom-agent configuration is available for only two patents, compared with seven for several other configurations, we treat this result as illustrative rather than as a fully balanced comparison.

Taken together, the results do not suggest that agentic structure replaces model or inference-time scaling. Rather, they indicate that \emph{model capability, inference-time reasoning, and domain-specific agentic scaffolding are complementary mechanisms for improving professional drafting performance}.

\subsection{Per-Dimension Model and Agent Results}

\begin{figure*}[t]
    \centering
    \includegraphics[width=\textwidth]
    {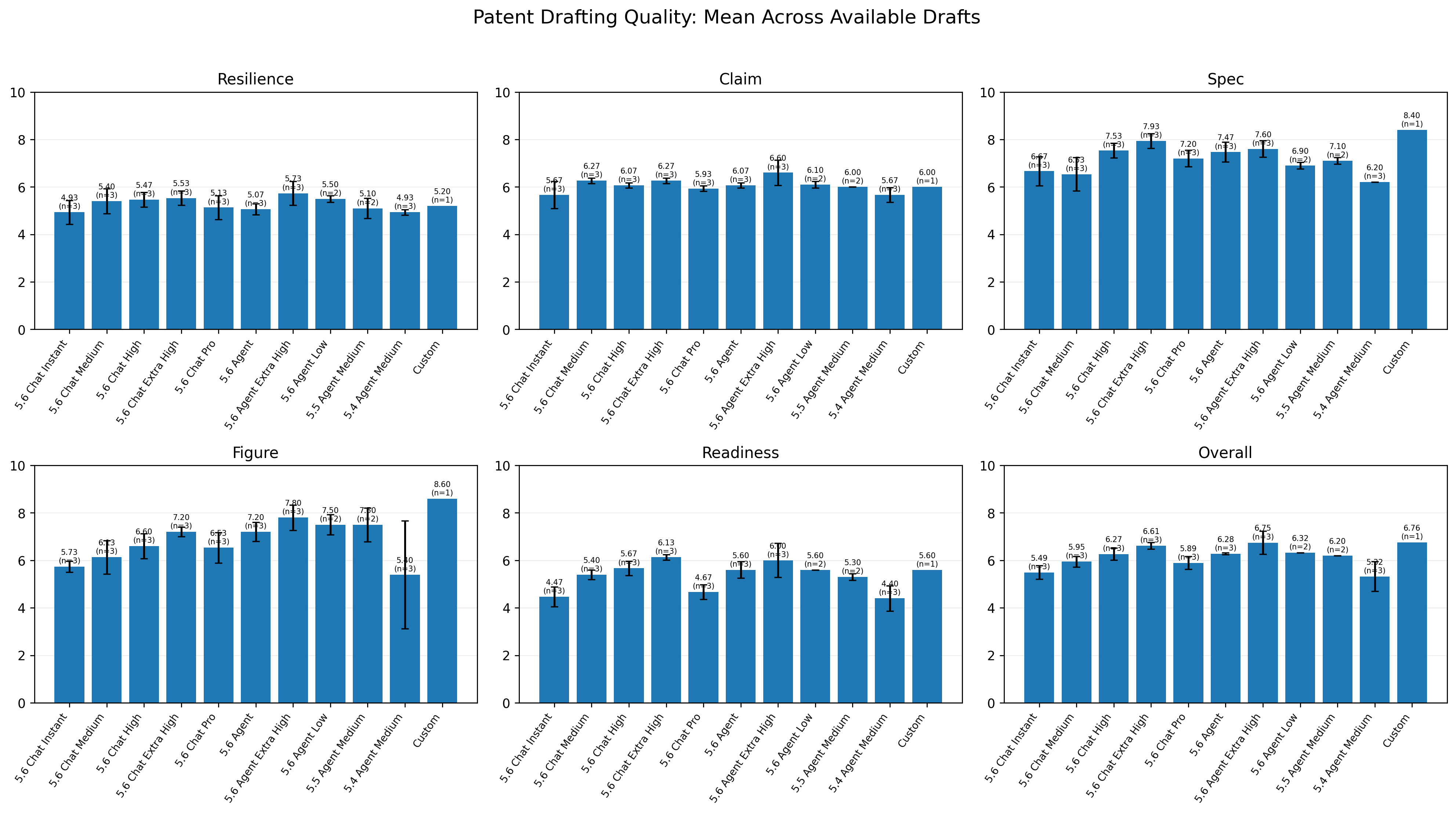}
    \caption{
    Extended quality comparison across model, reasoning-effort, and agentic configurations. Error bars summarize variability across the available patent drafts. 
    }
    \label{fig:extended-config}
\end{figure*}

Figure~\ref{fig:extended-config} reports the five individual quality dimensions and overall score for the evaluated model and agent configurations.

\subsection{Inference-Time Scaling}
\label{app:time-results}

\begin{figure}[t]
    \centering
    \includegraphics[width=\columnwidth]
    {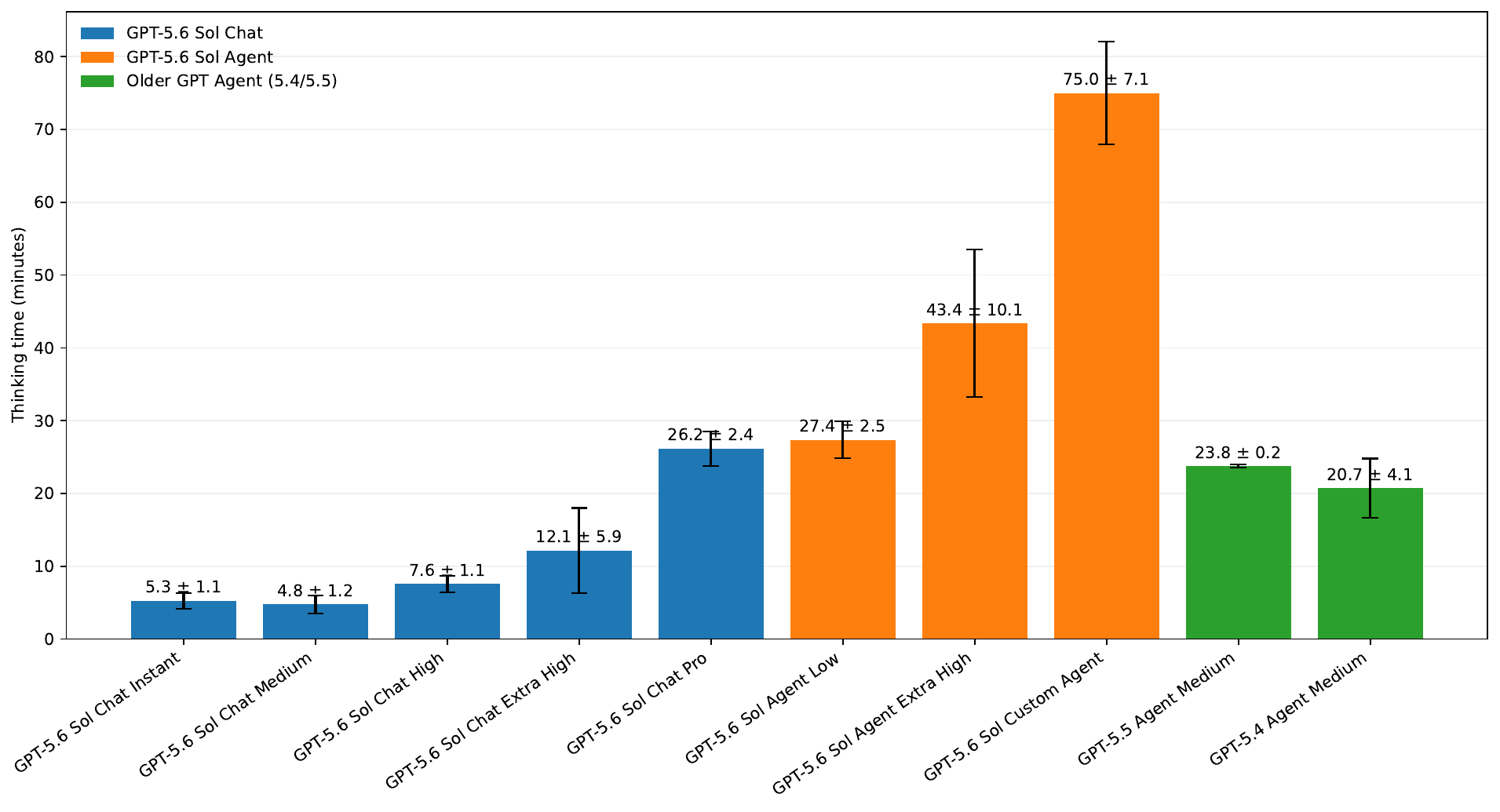}
    \caption{
    Mean patent-drafting thinking time for different model and agent configurations. Error bars indicate one standard deviation across available drafting runs.
    }
    \label{fig:thinking-time}
\end{figure}

We next examine the computational cost of the different configurations using measured end-to-end thinking time during patent generation. Figure~\ref{fig:thinking-time} reports mean generation time with one standard deviation over repeated patent-drafting runs.

Chat-mode inference is substantially faster than full agent execution. For example, mean drafting times are approximately $5.3$, $4.8$, $7.6$, and $12.1$ minutes for Instant, Medium, High, and Extra-High chat configurations, respectively. 
GPT-5.6 Pro requires approximately $26.2$ minutes, while the GPT-5.6 agent with Medium reasoning requires $28.3$ minutes and the Extra-High agent approximately $41.5$ minutes.

\subsection{Quality Versus Thinking Time by Dimension}

\begin{figure*}[t]
    \centering
    \includegraphics[width=\textwidth]
    {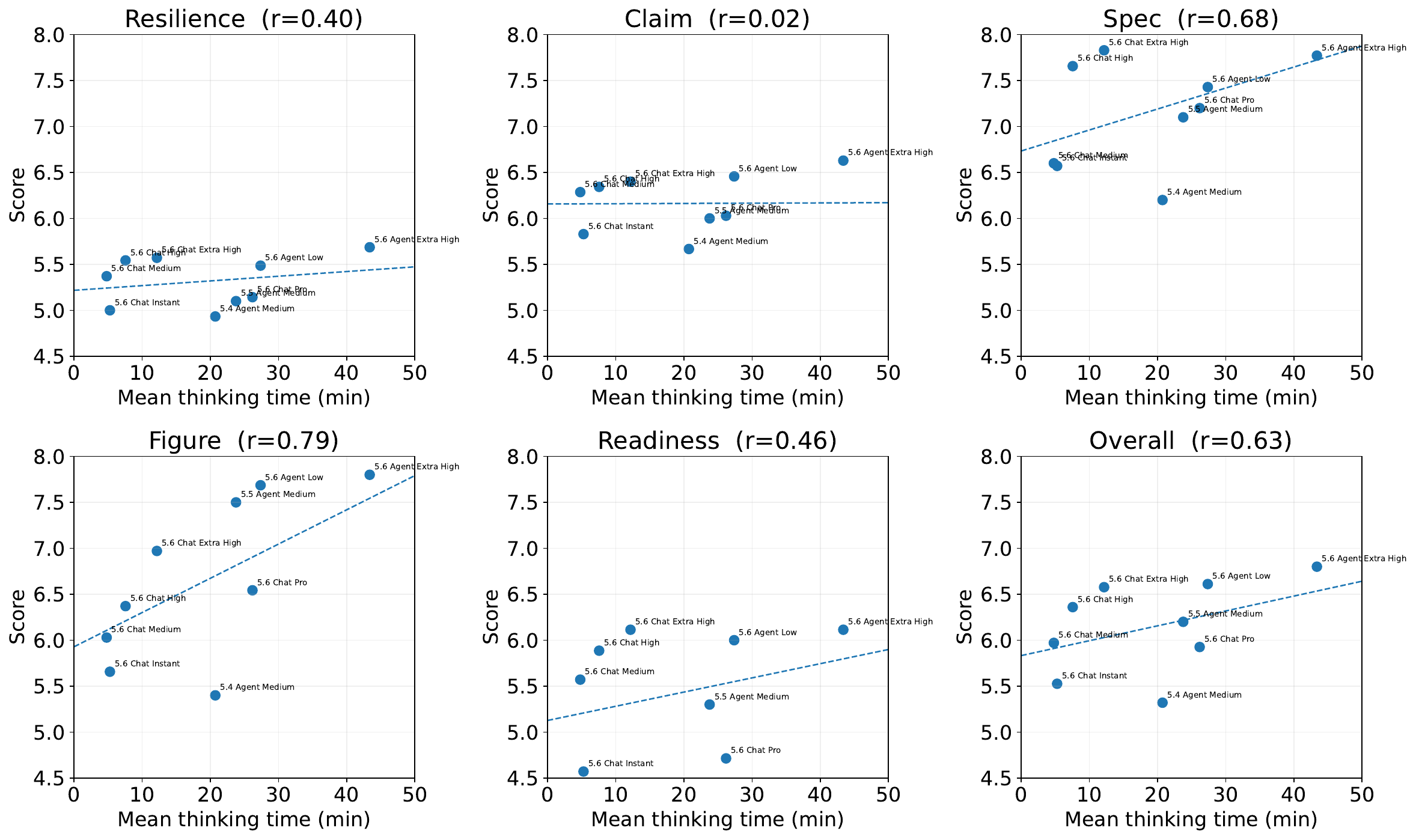}
    \caption{
    Patent-drafting quality versus mean thinking time for each of the five quality dimensions and the overall score. Each point represents one model/reasoning/agent configuration.
    }
    \label{fig:time-six}
\end{figure*}

The relationship between drafting time and quality differs across evaluation dimensions. Figure~\ref{fig:time-six} shows separate configuration-level scatter plots.

In our current configuration-level measurements, the Pearson correlations between thinking time and the five dimensions are approximately
\begin{align}
    r_{\mathrm{Resilience}} &= 0.40,\\
    r_{\mathrm{Claim}} &= 0.02,\\
    r_{\mathrm{Disclosure}} &= 0.68,\\
    r_{\mathrm{Figure}} &= 0.79,\\
    r_{\mathrm{Readiness}} &= 0.46,
\end{align}
with an overall correlation of approximately $r=0.63$. These values are descriptive because the number of configurations is small and the observations are not randomized allocations of inference compute.

\subsection{Per-Dimension QA-Guided Revision}

\begin{figure*}[t]
    \centering
    \includegraphics[width=\textwidth]
    {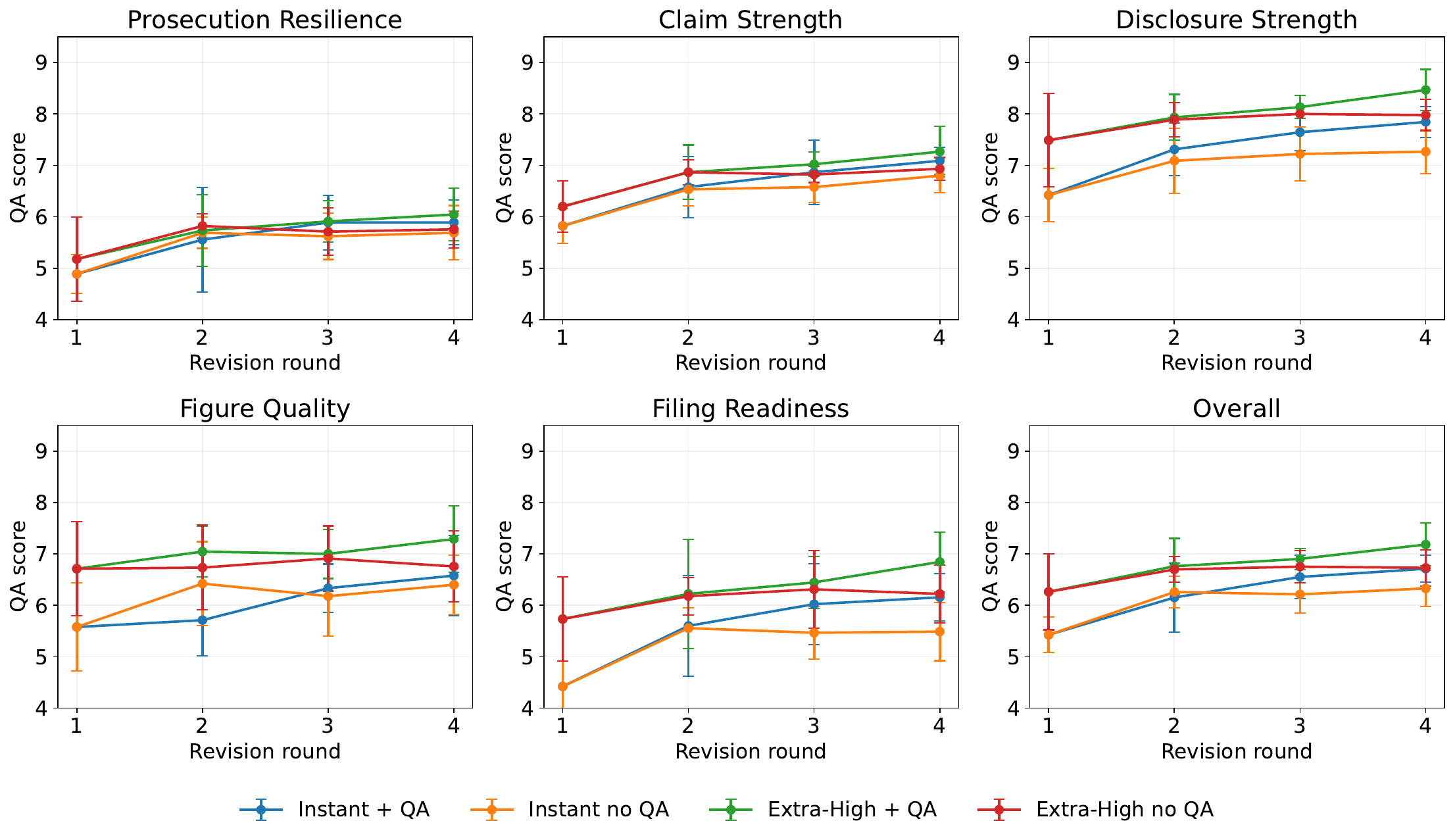}
    \caption{
     QA-guided revision trajectory for each patent-quality dimension and the overall score.
    }
    \label{fig:tunecomp-revision}
\end{figure*}

Figure~\ref{fig:tunecomp-revision} provides the full five-axis revision trajectories that complement the overall-score trends shown in the main paper.

\subsection{Thinking Time Over Revision}

\begin{figure}
    \centering
    \includegraphics[width=0.8\linewidth]{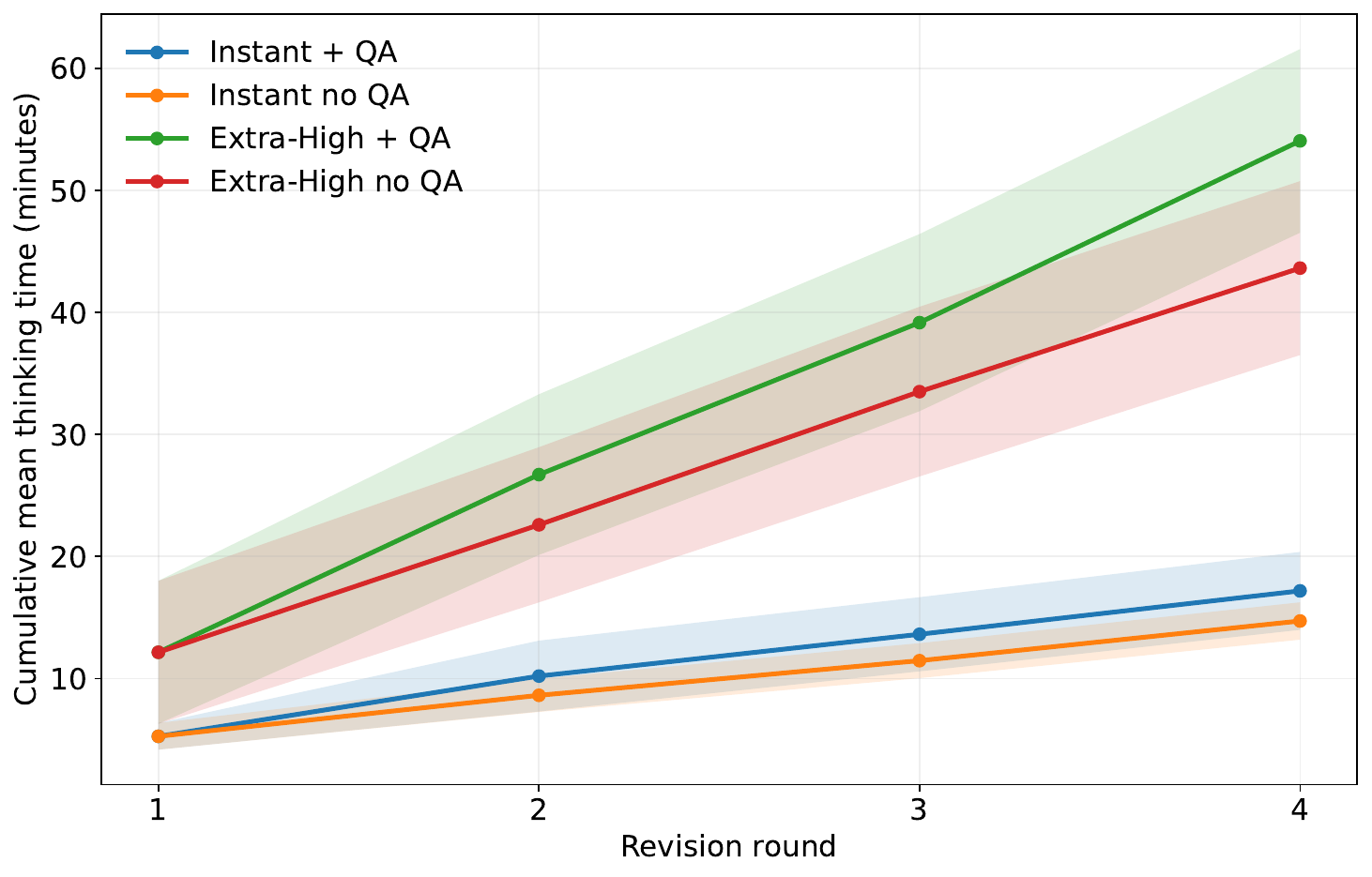}
    \caption{Thinking time across revisions: mean and standard deviation.}
    \label{fig:revision_thinking}
\end{figure}

Figure~\ref{fig:revision_thinking} shows the accumulated thinking time over patent revision with and without QA judge guidance. 
Instant reasoning is approximately 3.3-times faster than Extra-High reasoning.

\subsection{AI--Expert Agreement}

\begin{figure*}[t]
    \centering
    \includegraphics[width=\textwidth]
    {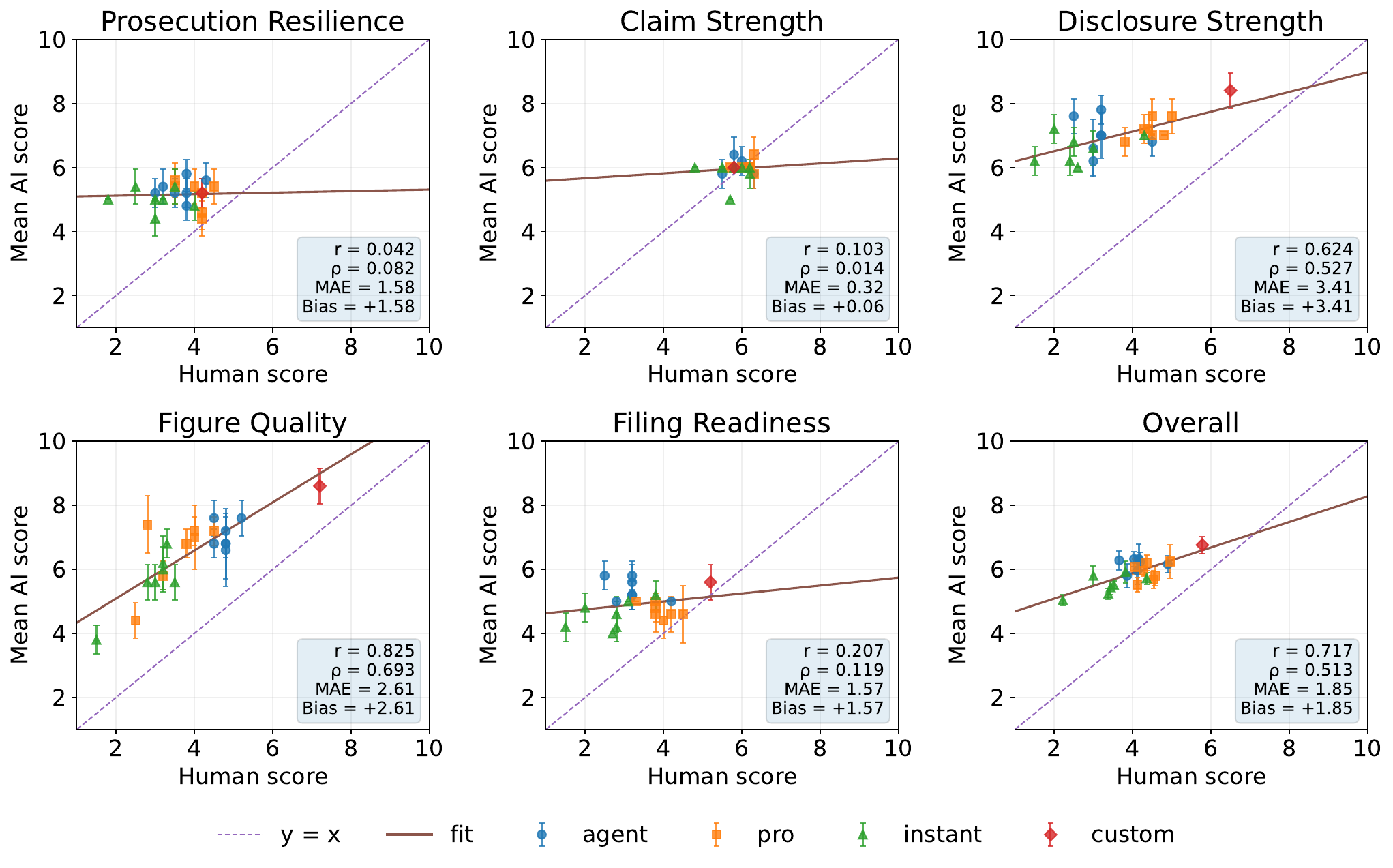}
    \caption{
    AI QA score versus professional patent-attorney score for the five quality dimensions and the overall score. AI scores are averaged over five independent QA runs.
    }
    \label{fig:expert-scatter-app}
\end{figure*}

\begin{figure}
    \centering
    \includegraphics[width=\linewidth]{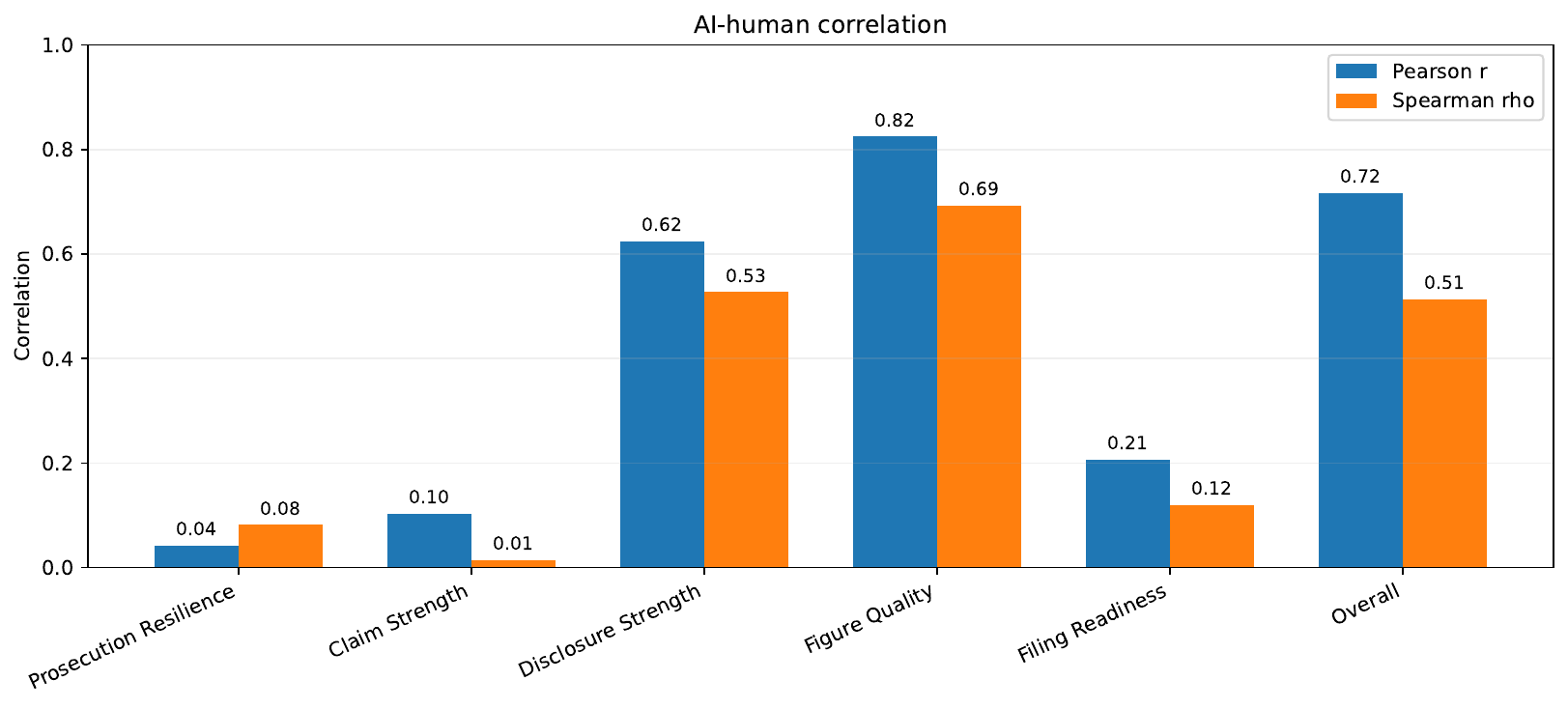}
    \caption{Correlation in QA evaluations between AI score and professional patent-attorney score.}
    \label{fig:correlation}
\end{figure}

Figure~\ref{fig:expert-scatter-app} shows the complete AI-versus-attorney comparison.
Figure~\ref{fig:correlation} plots the correlation between AI score and professional attorney score: Pearson $r$ and Spearman $\rho$.

The results demonstrate two different forms of agreement. Some dimensions show strong relative association but systematic absolute bias, whereas other dimensions exhibit similar absolute scale but weak discrimination across drafts. This motivates reporting both correlation and calibration-oriented metrics such as MAE and bias.

\subsection{QA Agent Repeatability}

Because each draft is evaluated five times, we can also characterize the stability of the AI evaluator. For draft $i$ and quality dimension $k$, we compute
\begin{equation}
    \sigma_{i,k}
    =
    \mathrm{Std}
    \left(
        q_{i,k}^{(1)},
        \ldots,
        q_{i,k}^{(5)}
    \right).
\end{equation}

Low $\sigma_{i,k}$ indicates that repeated QA executions return similar judgments, whereas high variability identifies dimensions or drafts for which the evaluator is less stable.

\begin{figure}[t]
    \centering
    \includegraphics[width=\columnwidth]
    {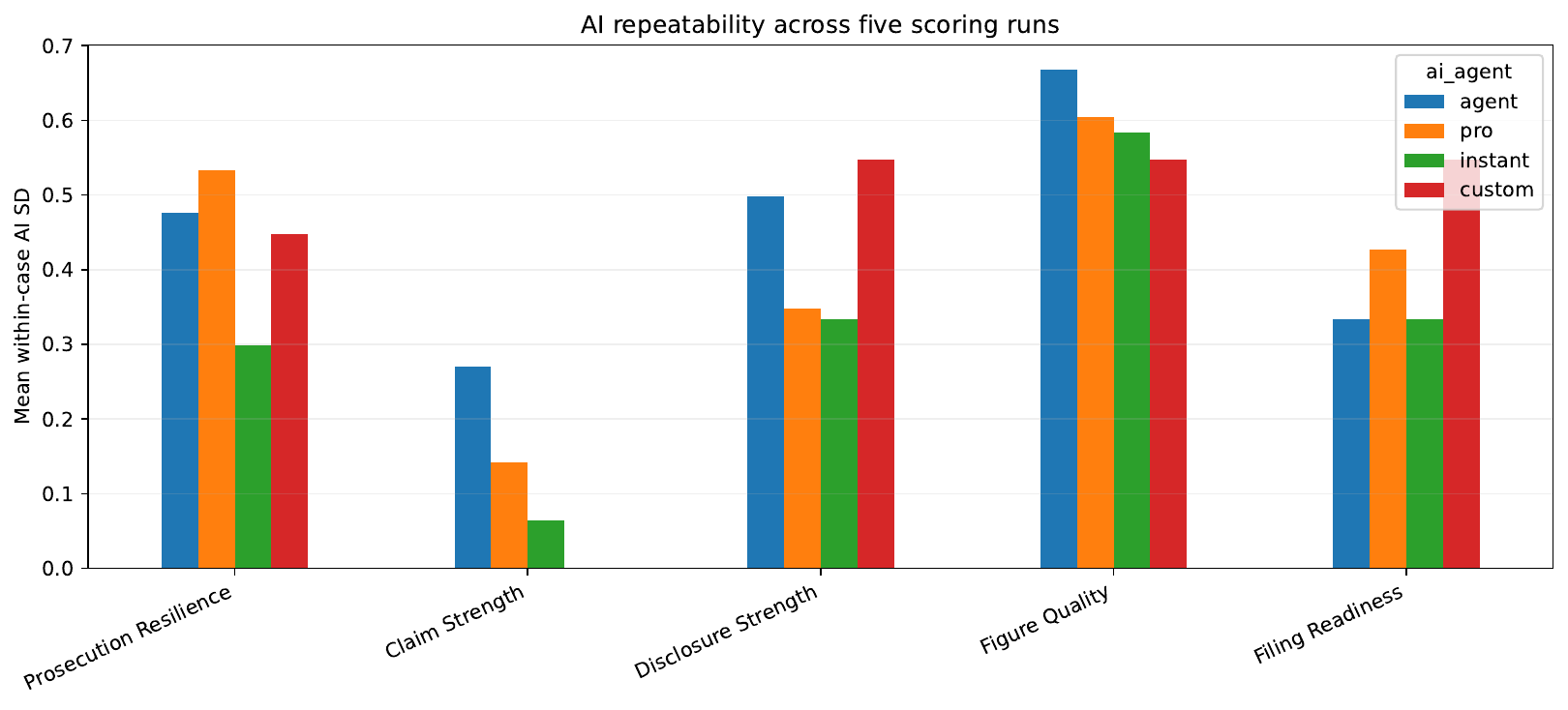}
    \caption{
    Run-to-run variability of repeated AI QA evaluation across patent-quality dimensions.
    }
    \label{fig:qa-repeatability}
\end{figure}

Figure~\ref{fig:qa-repeatability} shows the mean of within-class standard deviation across QA agent runs. 
We observe that the standard deviation is mostly less than 0.5 for integer scoring. 
It suggests that the QA agent gives relatively stable scores. 

\subsection{Summary of Findings}

Our experiments yield several complementary observations. First, judge-assessed patent quality generally improves with stronger model capability, greater inference-time reasoning, and domain-specific agentic scaffolding, although additional computation alone does not guarantee proportional gains. The skilled human drafter provides a separate generation baseline, distinct from the professional patent attorney used for expert evaluation.

Second, iterative revision substantially improves the LLM judge's scores. Generic revision provides considerable initial gains but tends to saturate, whereas QA-guided revision continues to improve through the evaluated revision rounds. This effect is particularly pronounced for the low-reasoning generator. After repeated judge-guided refinement, low-reasoning generation approaches the judge-assessed quality achieved by substantially more expensive high-reasoning generation, suggesting that revision depth and structured evaluator feedback can partially compensate for inference strength.

Third, comparison with the professional patent attorney shows that LLM-judge reliability is strongly dependent on the quality dimension. Some dimensions exhibit meaningful relative agreement with expert judgments despite systematic score bias, while others show weak ranking agreement. Conversely, good absolute calibration does not necessarily imply strong correlation with expert rankings. These results emphasize that correlation, calibration, and repeatability capture distinct properties of an automated judge.

Finally, the revision experiment illustrates an important limitation of judge-in-the-loop optimization. Increasing scores under repeated optimization demonstrates that the judge provides an actionable optimization signal, but does not by itself establish corresponding improvement under professional expert evaluation. Together, our findings suggest that LLM judges can be useful components of professional agentic workflows, while independent expert validation remains important when interpreting judge-optimized performance.

\section{Limitations and Responsible Use}
\label{app:limitations}

\subsection{Experimental Limitations}

Our experiments are intended as an initial controlled study of professional agentic drafting and have several limitations.

First, the number of technical inventions is modest, and not every model or agent configuration was evaluated on every patent. We therefore distinguish balanced evaluations from illustrative results and explicitly report the number of contributing cases where appropriate.

Second, expert human evaluation is currently based on a single professional patent attorney. Patent quality involves subjective professional judgment, and future evaluation should include multiple practitioners to characterize inter-expert variability and establish an appropriate human performance range.

Third, the underlying frontier models and their reasoning modes are rapidly evolving. Configuration names such as Instant, High, Extra High, and Pro identify product-level inference settings rather than stable computational algorithms. The precise behavior, resource allocation, and latency of such systems may change over time.

Fourth, the automatic QA evaluator is itself an LLM-based system. Although we compare its scores with professional evaluation, it exhibits dimension-dependent calibration bias and should not be interpreted as an objective ground-truth metric.

Finally, our present evaluation measures draft quality before prosecution. Long-term patent value depends on factors that cannot be established from an initial application alone, including examiner search results, prosecution history, allowed claim scope, enforceability, litigation outcomes, and commercial relevance.

\subsection{Professional Oversight}

Vibe Patenting is intended to automate substantial portions of \emph{patent production}, not to eliminate professional judgment from the intellectual-property process. Important decisions remain outside the scope of autonomous drafting, including:
\begin{itemize}
    \item determining whether an invention should be patented;
    \item verifying technical accuracy with the inventors;
    \item determining inventorship;
    \item assessing legal obligations and disclosure requirements;
    \item making final patentability and claim-strategy judgments;
    \item approving the application for filing; and
    \item conducting prosecution before a patent office.
\end{itemize}

A useful operating model is therefore
\begin{equation}
    \text{Researchers create and validate}
    \rightarrow
    \text{AI structures and drafts}
    \rightarrow
    \text{Professionals review and approve}.
\end{equation}

The objective is to reduce repetitive drafting and coordination effort so that researchers and patent practitioners can spend more time on technical truth, strategic judgment, and portfolio decisions.

\subsection{Confidentiality and Data Governance}

Patent drafting frequently involves unpublished and commercially sensitive technical information. Deployment of an automated patent-drafting system therefore requires appropriate controls over data retention, model access, logging, external tool usage, and disclosure of confidential materials. Organizations should ensure that the AI infrastructure used for drafting is compatible with their confidentiality, security, and intellectual-property policies.

\subsection{Future Work}

Several directions follow naturally from the present study:
\begin{itemize}
    \item evaluation on a substantially larger and more diverse collection of inventions;
    \item evaluation by multiple patent practitioners and measurement of human--human agreement;
    \item comparison with commercial patent-drafting systems where reproducible access is available;
    \item controlled ablation of individual professional skills and PISE components;
    \item evaluation using open-weight models and reproducible inference budgets;
    \item optimization of the quality--compute trade-off through adaptive reasoning allocation;
    \item learned stopping criteria for QA-guided revision;
    \item longitudinal evaluation using prosecution outcomes and allowed claim scope; and
    \item extension of structured professional-agent workflows to other scientific, engineering, legal, and business tasks.
\end{itemize}

\end{document}